\documentclass[sigconf]{acmart}

\usepackage{bm}
\usepackage{wrapfig}
\usepackage{amsmath}

\usepackage{amssymb}
\usepackage{mathtools}
\usepackage{amsthm}
\usepackage{makecell}
\usepackage{subfigure}
\usepackage{enumitem,multirow}
\usepackage{algorithm}
\usepackage[noend]{algorithmic}
\usepackage[capitalize,noabbrev]{cleveref}
\usepackage{subcaption} 

\AtBeginDocument{%
  }

\setcopyright{acmlicensed}
\copyrightyear{2027}
\acmDOI{XXXXXXX.XXXXXXX}

\acmISBN{978-1-4503-XXXX-X/2018/06}

\begin{document}

\title{KO: Kinetics-inspired Neural Optimizer with PDE Simulation Approaches}

\newcommand{\sjtuaffiliation}{%
  \affiliation{%
    \institution{Shanghai Jiao Tong University}
    \city{Shanghai}
    \country{China}
  }
}

\newcommand{\cmuaffiliation}{%
  \affiliation{%
    \institution{Carnegie Mellon University}
    \city{Pittsburgh}
    \state{Pennsylvania}
    \country{US}
  }
}

\author{Mingquan Feng}
\authornote{Both authors contributed equally to this research.}
\sjtuaffiliation

\author{Yixin Huang}
\authornotemark[1]
\cmuaffiliation

\author{Yifan Fu}
\sjtuaffiliation

\author{Shaobo Wang}
\sjtuaffiliation

\author{Junchi Yan}
\sjtuaffiliation
\email{yanjunchi@sjtu.edu.cn}

\renewcommand{\shortauthors}{Trovato et al.}

\begin{abstract}
The design of effective optimization algorithms for neural networks remains a fundamental challenge, and most existing methods rely on heuristic extensions of gradient-based updates. We introduce KO (Kinetics-inspired Optimizer), a plug-and-play optimization module grounded in kinetic theory and partial differential equations. KO models parameter dynamics as a particle system, augmenting standard gradient updates with stochastic interactions induced by a discretization of the Boltzmann transport equation. This mechanism naturally promotes parameter diversity and mitigates weight condensation, the tendency of parameters to collapse into low-dimensional subspaces, a phenomenon closely associated with degraded generalization. We provide both a rigorous theoretical analysis and a physical interpretation, showing that KO provably increases parameter diversity while preserving convergence guarantees. Extensive experiments on image classification benchmarks (CIFAR-10/100, ImageNet) and large-scale language model pretraining demonstrate that KO consistently improves accuracy over competitive baselines with negligible additional computational cost.
\end{abstract}


\begin{teaserfigure}
\centering
 \includegraphics[width=0.9\textwidth]{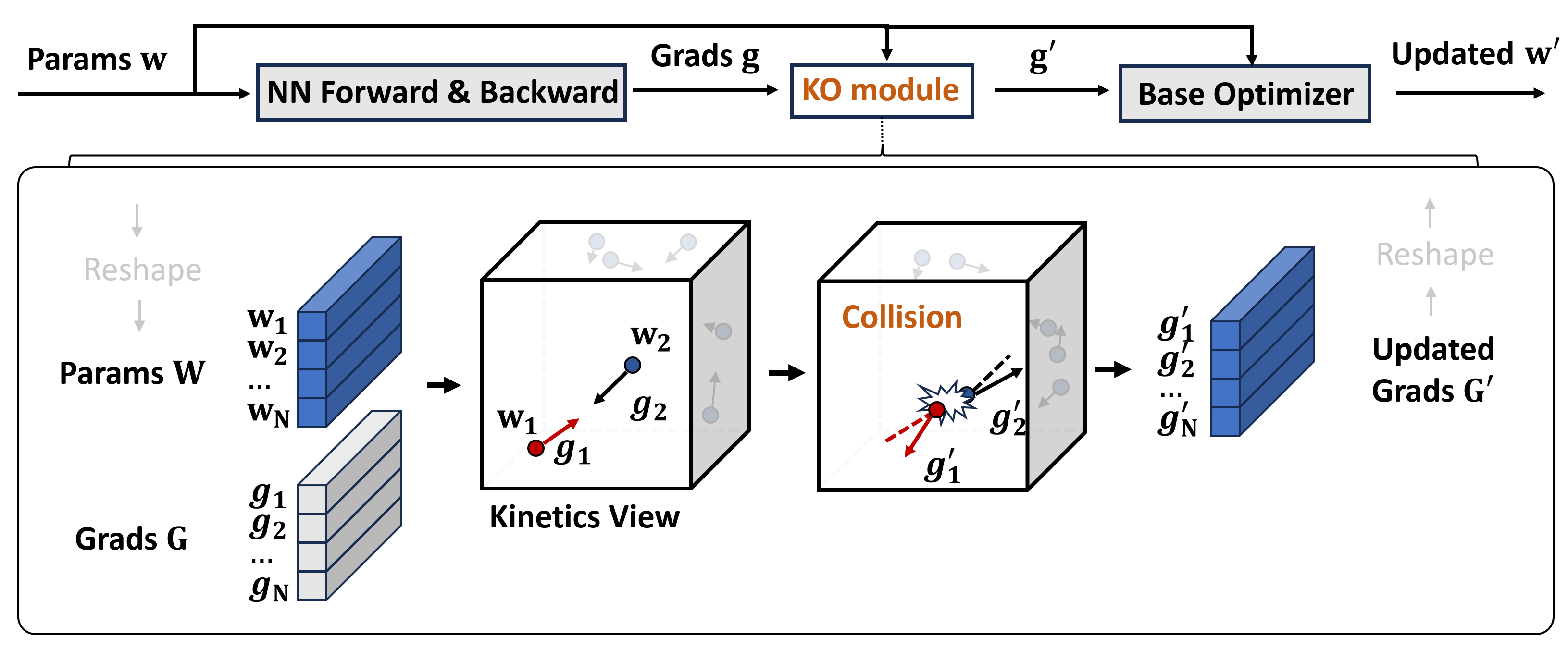}
  \caption{The architecture of KO. The upper part depicts the workflow: the gradient is first calculated by backpropagation, then refined by the kinetic module, and finally fed into the base optimizer. The lower part details the kinetic module. Weights and gradients are viewed as positions and velocities of particles; upon collision, particle velocities are updated following the collision model.}
  \label{fig:ki-opt}
\end{teaserfigure}


\maketitle

\section{Introduction}
Optimization algorithms are central to training neural networks, shaping convergence, generalization, and robustness. Gradient-based methods, including SGD~\citep{robbins1951stochastic} and adaptive optimizers such as Adam~\citep{kingma2014adam} and AdamW~\citep{loshchilov2017decoupled}, dominate in practice due to their simplicity and empirical effectiveness~\citep{devlin2019bertpretrainingdeepbidirectional, bai2023qwentechnicalreport, deepseekai2025deepseekv3technicalreport}. Recent advances—such as SAM~\citep{foret2021sharpnessawareminimizationefficientlyimproving}, Lion~\citep{chen2023symbolic}, and Sophia~\citep{liu2024sophia}—incorporate sharpness-aware, sign-based, or second-order information. In parallel, physics-inspired approaches interpret training as dynamical systems, including electromagnetic analogies~\citep{abe2016electromagnetic}.

Despite their success, existing optimizers lack mechanisms to regulate the distribution of network parameters during training. Empirical studies~\citep{zhou2022towards, chen2023phase} show that parameters often collapse into low-dimensional subspaces, a phenomenon known as weight condensation~\citep{xu2025overview}. Formally, weight condensation refers to the tendency of weight vectors within a layer to become highly correlated, as measured by the average pairwise absolute cosine similarity:
\begin{equation}
    \rho(\mathbf{w}) = \frac{1}{N (N- 1)} \sum_{i=1,j=1, i\ne j}^{N} \frac{|\mathbf{w}_{i}^\top \mathbf{w}_{j}|}{\Vert \mathbf{w}_{i}\Vert_2 \Vert \mathbf{w}_{j} \Vert_2},
    \label{eq: wc_def}
\end{equation}
where $\mathbf{w}_i$ denotes the $i$-th neuron's weight vector in a layer with $N$ neurons. High condensation implies reduced effective model capacity and has been shown to degrade generalizability~\citep{jin2020doesweightcorrelationaffect}. Although regularization techniques such as weight decay or dropout partially slow condensation \citep{zhang2024implicit}, they operate by modifying the loss landscape rather than intervening directly in the optimization dynamics, and we show empirically that weight decay does not prevent ultimate condensation at convergence (see Appendix~\ref{appx:condense}). Furthermore, current physics-inspired methods often oversimplify the analogy between physical systems and neural networks, neglecting the stochastic, many-body interactions that govern parameter evolution. This gap raises a crucial question: \emph{Can we design optimizers that inherently promote parameter diversity by simulating the microscopic collision-driven dynamics observed in kinetic systems?}

We introduce KO, a kinetics-inspired neural optimizer that models parameter training as the evolution of a particle system. By drawing an analogy between the network parameters and colliding particles, KO uses a discretized Boltzmann transport equation (BTE)~\citep{boltzmann2015relationship} to capture stochastic interactions. Specifically, neuron weights are treated as particle positions and gradients as velocities, with collisions applied at the gradient level to separate neurons that are close in weight space and aligned in direction. This mechanism promotes parameter diversity, counteracting weight condensation while preserving the behavior of the base optimizer. Our main contributions are as follows.

\begin{enumerate}[noitemsep,topsep=0pt,leftmargin=*,itemsep=2pt]
    \item \textbf{Physics-driven optimization framework:} We propose a novel optimizer gadget that integrates kinetic theory and PDE simulation concepts into neural network training. Two collision variants are introduced---hard collision based on Direct Simulation Monte Carlo, and a computationally efficient soft collision derived from the same physical principles. 
    \item \textbf{Theoretical analysis:} We prove that both collision mechanisms provably reduce weight correlation under mild assumptions. We further establish convergence guarantees showing that the collision perturbations, while structured, do not compromise the convergence properties of the base optimizer. 
    \item \textbf{Empirical validation:} Extensive experiments across image classification (CIFAR-10/100, ImageNet) and large-scale language model pretraining (Qwen2-0.5B/1.5B) demonstrate that KO consistently improves accuracy over competitive baselines.
\end{enumerate}

\section{Related Work}
\textbf{Deep-learning Gradient-based Optimizers.}
Advances in optimization algorithms have been central to deep learning. Early methods such as SGD~\citep{ruder2016overview} and momentum~\cite{nesterov1983method} established gradient-based updates with improved convergence. Adaptive methods, including AdaGrad~\citep{duchi2011adaptive}, RMSProp~\citep{RMSProp}, and Adam~\citep{kingma2014adam}, further adjust learning rates based on gradient history. Second-order approaches (e.g., BFGS~\citep{fletcher2000practical}, L-BFGS~\citep{liu1989limited}) improve convergence at higher computational cost. More recent methods focus on scalability and generalization, such as LAMB~\citep{you2019large}, AdaFactor~\citep{shazeer2018adafactor}, SAM~\citep{foret2021sharpnessawareminimizationefficientlyimproving}, Lion~\citep{chen2023symbolic}, and Sophia~\citep{liu2024sophia}. Our method is orthogonal to these approaches: KO modifies gradients directly and can be combined with existing optimizers to enhance parameter diversity. While gradient noise has been explored for regularization~\citep{neelakantan2015addinggradientnoiseimproves}, KO introduces structured perturbations that explicitly reduce weight correlation, rather than isotropic noise.

\textbf{Physics-Inspired Metaheuristics Optimizers.}
Researchers have taken inspiration from non-linear physical phenomena to formulate meta-heuristic optimization algorithms. For example, gravitational search algorithms \citep{formato2007central, rashedi2009gsa} model search agents as masses interacting according to Newtonian gravity. Similarly, electromagnetism-based algorithms \citep{javidy2015ions, abe2016electromagnetic, yadav2019aefa} define agents as charged particles driven by electrostatic or magnetic forces. Other metaheuristics draw from fluid mechanics \citep{dougan2015new, tahani2019flow}, optics \citep{kaveh2012new, kashan2015new}, thermodynamics \citep{moein2014kgmo, kaveh2017novel}, and nuclear physics \citep{wei2019nuclear}. However, these are gradient-free metaheuristics and are not directly applicable to deep learning optimization, lacking a clear connection between physics and gradient-based network dynamics.

A related line of work studies physics-inspired optimization via mean-field frameworks. \cite{Mei_2018} characterizes SGD through PDE-based distributional dynamics, while \cite{Rotskoff_2022} models particle interactions using sparse, potential-based formulations. Building on the latter, we instead model interactions via \emph{collisions} rather than continuous potentials. Moreover, unlike mean-field approaches that assume the infinite-particle limit, we operate in the finite-neuron regime, where Direct Simulation Monte Carlo is more appropriate. Further discussion is provided in Appendix~\ref{appx: conn}.

\textbf{Similarity in Neural Networks.} Similarity has been widely studied in machine learning. For instance, \cite{andriopoulos2024prevalenceneuralcollapseneural} analyzes Neural Collapse, where features of the same class converge to their class mean, focusing on activation-space geometry. In contrast, our work studies \emph{weight-space} similarity, which is input-independent and concerns the structural diversity of neurons. The two are orthogonal: a model may exhibit neural collapse while maintaining diverse weights, or vice versa. \cite{jin2020doesweightcorrelationaffect} examines the impact of weight similarity by introducing correlation-based regularization. In contrast, we address weight condensation at the optimization level by modifying gradient updates to induce repulsive forces between similar neurons. This avoids the trade-off between task loss and regularization inherent in penalty-based methods. We further provide theoretical guarantees on reducing weight correlation while preserving convergence, and include empirical comparisons in Section~\ref{sec:experiments}.

\section{Preliminary}
\subsection{Kinetic Theory}
The kinetic molecular theory of ideal gases~\citep{loeb2004kinetic} is built on several simplifying assumptions. First, a gas is composed of identical particles called molecules. Second, these molecules are in constant motion and follow Newton’s laws of motion. Third, the molecules act as small, elastic spheres with negligible volume. This implies that the collisions within the system are perfectly energy-conserving. Finally, interactions are limited to collisions, with no long-range attractive or repulsive forces.

While these assumptions provide a useful microscopic picture, directly tracking the motion of individual particles becomes intractable in systems with a huge number of molecules. To address this, one instead turns to the statistical distributions to describe the system’s dynamics. The change in the number of particles $N$ is hence characterized by $f$, a density function defined in a 7-dimensional phase space: $\text{d} N = f(\mathbf{x}, \mathbf{p}, t) \, \text{d}^3 \mathbf{x} \, \text{d}^3\mathbf{p}$. If position $\mathbf{x}$ and momentum $\mathbf{p}$ evolve according to the Hamiltonian equations under an external force $F_{ex}$ on the system, then the density function $f$ satisfies the Boltzmann Transport Equation (BTE): $\frac{\partial f}{\partial t} + \frac{\mathbf{p}}{m} \cdot \nabla_\mathbf{x} f + F_{ex} \cdot \nabla_\mathbf{p} f = \left( \frac{\partial f}{\partial t} \right)_{coll}$, where the right-hand term captures the changes in distribution caused by particle collisions. Since no closed-form expression exists for this term, it is typically approximated empirically. As a partial differential equation, BTE governs the temporal evolution of the distribution function $f$. The Direct Simulation Monte Carlo (DSMC) method~\citep{ivanov1988analysis} and the lattice Boltzmann method~\citep{kruger2017lattice} are two common numerical approaches for solving BTE.
\subsection{Direct Simulation Monte Carlo (DSMC)}
The DSMC is a stochastic approach that solves BTE for dilute gases by simulating particle motion. In DSMC, each simulation particle represents $F_{N}$ physical molecules. The method discretizes the computational domain into cells and then tracks the evolution of the particles through drift, wall collision, and particle collision in each cell.

During the drift phase, particles follow straight-line trajectories determined by their current velocities, with the assumption that no collisions occur during this phase. When particles hit domain boundaries, the wall collision step processes these interactions according to the boundary conditions. These two steps provide deterministic updates.  

Stochasticity is introduced in the particle collision step. Particle collision occurs only between particles that fall into the same cell. The hard sphere model, where the collision probability is proportional to the relative velocity of the particle pairs, is utilized to model the collision:
\begin{equation}
    P_{\mathrm {coll} }[i,j]=\frac{|\mathbf {v} _{i}-\mathbf {v} _{j}|} {\sum _{m=1}^{N_{\mathrm {c} }}\sum _{n=1}^{m-1}|\mathbf {v} _{m}-\mathbf {v} _{n}|},
    \label{eq:prob}
\end{equation}
where $N_{\mathrm {c}}$ is the number of particles in the cell and $\mathbf {v}$ is the particle velocity. The DSMC method utilizes a rejection sampling approach to efficiently approximate collision probabilities, as direct implementation of Eq. \ref{eq:prob} would be computationally prohibitive. The algorithm begins by estimating potential collision pairs $M_{\mathrm {cand} }$ using the no-time counter technique: $ M_{\mathrm {cand} }=\frac{N_{\mathrm {c} }(N_{\mathrm {c} }-1)F_{N}\pi d^{2}v_{\mathrm {r} }^{\max }\tau }{2V_{\mathrm {c} }}$, where $d$ is the diameter of the particle, $v_{\mathrm {r} }^{\max }$ is the maximum relative velocity, $\tau$ is the time step and $V_{\mathrm {c} }$ is the cell volume. Given this, we can randomly select $M_{\mathrm {cand} }$ pairs of candidate particles and determine whether the collision will occur given the following criterion: 
\begin{equation}
\begin{aligned}
|\mathbf {v} _{i}-\mathbf {v} _{j}|/v_{\mathrm {r} }^{\max } > \Re_1.
\end{aligned}
\label{eq: DSMC_accept}
\end{equation}
where $\Re_1$ is sampled from the uniform distribution $U(0,1)$. For accepted collisions, particle velocities are modified according to the chosen collision model without updating the spatial positions of the particles. This cell-wise procedure iterates until all potential collisions are processed.

The simulation employs the hard sphere collision model, which approximates particles as perfectly rigid bodies. This model strictly conserves both momentum and kinetic energy during collisions while randomly adjusting the direction of the scattering. Post-collision relative velocities are expressed in the polar coordinate form:
\begin{equation}
\begin{aligned}
\mathbf {v} _{\mathrm {r} }^{*}=v_{\mathrm {r} }[(\sin \theta \cos \phi ){\hat {\mathbf {x} }}+(\sin \theta \sin \phi ){\hat {\mathbf {y} }}+\cos \theta \,{\hat {\mathbf {z} }}],
\end{aligned}
\label{eq: DSMC_vr}
\end{equation}
where $\phi =2\pi \Re _{2}$, $\theta =\cos ^{-1}(2\Re _{3}-1)$ and $\Re _{2}$ and $\Re _{3}$ are numbers randomly sampled from the uniform distribution $U(0,1)$. If we further denote the center of mass velocity as $\mathbf {v} _{\mathrm {cm} }=(\mathbf {v} _{i}+\mathbf {v} _{j})/2$, the post-collision velocity can be calculated as:
\begin{equation}
\mathbf {v} _{i}^{*}=\mathbf {v} _{\mathrm {cm} }+\mathbf {v} _{\mathrm {r} }^{*}/2,
\mathbf {v} _{j}^{*}=\mathbf {v} _{\mathrm {cm} }-\mathbf {v} _{\mathrm {r} }^{*}/2,
    \label{eq: DSMC_deltav}
\end{equation}

\section{Methodology}
\subsection{Kinetics-inspired Neural Optimizer}
The intuition of the kinetics-inspired neural optimizer is to hinder a phenomenon called neuron condensation of neural network parameters~\citep{zhou2022towards}, where neurons in the same layer cluster together during training. The cosine similarity within the network layer is a key metric in evaluating the degree of condensation. We prove in Section \ref{section: understand} that excessive condensation obstructs model performance.

The proposed optimizer architecture is visualized in Fig. \ref{fig:ki-opt}. It is built on a base optimizer, such as SGD or Adam. We draw an analogy between gradient-based parameter updates in optimization and velocity-driven position updates in particle systems, which is detailed in Appendix~\ref{appx: conn}. As discussed earlier, optimization often leads to the formation of parameter clusters. To counter this effect, we introduce collisions at the gradient level to promote separation among neurons. Then the optimizer updates the parameters according to the base optimizer's rules. It is noteworthy that we operate at the gradient level rather than directly modifying parameter updates. This allows us to retain the core behavior of the base optimizer. This design makes our method a modular enhancement: it can be integrated into diverse optimizers without undermining their intrinsic properties. 

At a high level, the collisions can be seen as introducing physics-inspired perturbations to the gradients. However, unlike Gaussian noise, which is directionless and purely stochastic, our perturbations are theoretically motivated by the goal of alleviating weight condensation and enhancing generalization. We can even provide a mathematical proof of the effects of this directional mechanism. Thus, while both approaches involve gradient perturbation, the underlying principles and effects of our method are fundamentally different. In the following, we will present two types of collision formulation. 

\textbf{Hard Collision.} The hard-body collision is an ideal operation to separate neurons. It updates the gradients by simulating the hard-body collision using the DSMC method. 

First, we calculate the relative distance and velocity between neurons, along with the position and velocity of their center of mass, transforming the system to be center-of-mass:
\begin{equation}
\begin{aligned}
(\mathbf{w}_r)_{i,j}=|\mathbf {w} _{i}-\mathbf {w} _{j}|, \quad 
(\mathbf{g}_r)_{i,j}=|\mathbf {g} _{i}-\mathbf {g} _{j}|, \\
(\mathbf{w}_{\mathrm{cm}})_{i,j}=\frac{1}{2}(\mathbf {w} _{i}+\mathbf {w} _{j}), \quad (\mathbf{g}_{\mathrm{cm}})_{i,j}=\frac{1}{2}(\mathbf {g} _{i}+\mathbf {g} _{j}).
\end{aligned}
\label{eq: xv_r}
\end{equation}    
Secondly, we derive the formula for the changes of gradients to be applied after the collision, i.e., $\Delta \mathbf{g}$:
\begin{equation}
\begin{aligned}
    (\Delta \mathbf{g})_{i,j} = (\mathbf{g}_{\mathrm{cm}})_{i,j} + \frac{1}{2}(\mathbf{g}_r)_{i,j} \mathbf {n}_{i,j} - \mathbf {g}_{i}, \\
    (\Delta \mathbf{g})_{j,i} = (\mathbf{g}_{\mathrm{cm}})_{j,i} + \frac{1}{2}(\mathbf{g}_r)_{j,i} \mathbf {n}_{j,i} - \mathbf {g}_{j},
\end{aligned}
\label{eq: delta_v}
\end{equation}
where $\mathbf{n}_{i,j}$ is a unit vector sampled uniformly from the sphere and is not correlated with the weight matrix $\mathbf{w}$. We further require $\mathbf{n}_{j,i} = -\mathbf{n}_{i,j}$ to ensure zero-mean collision. This formulation generalizes Eq.~\ref{eq: DSMC_deltav}, where $(\mathbf{w}_r)_{i,j}\mathbf{n}_{i,j}$ and $(\mathbf{w}_r)_{j,i}\mathbf{n}_{j,i}$ originate from Eq.~\ref{eq: DSMC_vr} and represent the relative receding velocity after collision in the center-of-mass frame.
    
Thirdly, a collision hyperparameter $\text{coll}\_{\textrm{coef}}$ is introduced to control the collision. Unlike traditional DSMC, where collision only occurs between particles in the same cell, our method allows collisions between any pair of particles. For each pair of neurons $i,j$, the collision occurs if:
\begin{equation}
    \frac{(\mathbf{g}_r)_{i,j} \cdot (U_r)_{i,j}}{g_r^{\mathrm{max}}}>1-\text{coll}\_{\textrm{coef}},
    \label{eq: coll_mask}
\end{equation}
where $(U_r)_{i,j} = e^{-(\mathbf{w}_r)_{i,j}}$, $g_r^{\mathrm{max}} = \max(\mathbf{g}_r)$. This equation is adapted from the original collision filter in Eq. \ref{eq: DSMC_accept}. $(U_r)_{i,j}$ is introduced as a soft locality constraint to reduce the rate of semantically meaningless collisions. To better control the collision percentage, we further introduce $U_r$ to reflect the effective mean free distance. As the relative distance $(\mathbf{w}_r)_{i,j}$ increases, $(U_r)_{i,j}$ decreases, reducing the probability of the collision between neurons $i$ and $j$. 

Finally, we update the gradients of the neurons by the aforementioned collision terms. The update algorithm is summarized in Algorithm \ref{alg:hard}.
\begin{equation}
\begin{aligned}
\mathbf {g} _{i}^{*} &=\mathbf {g} _{i} + \sum_{j \text{ in accepted pair }i,j } (\Delta \mathbf{g})_{i,j}, \\
\end{aligned}
\label{eq: xv_new}
\end{equation}

\begin{algorithm}[tb]
\caption{Hard-Collision-Based Gradient Updates}
\label{alg:hard}
\begin{algorithmic}[1]
\STATE \textbf{Input:} layer weight $\mathbf{w}\in \mathbb{R}^{N\times D}$, layer gradient $\mathbf{g}\in \mathbb{R}^{N\times D}$, hyper-parameters: $\text{coll}\_{\textrm{coef}}$.
\STATE \textbf{Output:} updated layer gradient $\mathbf{g}\in \mathbb{R}^{N\times D}$.
\STATE Calculate relative properties $\mathbf{w}_r, \mathbf{g}_r$ and center-of-mass properties $\mathbf{w}_{\mathrm{cm}}, \mathbf{g}_{\mathrm{cm}}$ by Eq.~\ref{eq: xv_r};
\STATE Calculate the full velocity change $\Delta \mathbf{g}$ by Eq.~\ref{eq: delta_v};
\STATE Select collision pairs by Eq.~\ref{eq: coll_mask};
\STATE Apply velocity and position change by Eq.~\ref{eq: xv_new}, get new gradient $\mathbf{g}$;
\STATE \textbf{Return} $\mathbf{g}$;
\end{algorithmic}
\end{algorithm}

\textbf{Soft Collision.} The aim of the kinetics-inspired neural optimizer is to reduce the network layers' weight similarity. In addition to directly simulating the collision during backpropagation, we can design a deterministic repulsion force to achieve similar effects. Unlike prior work that primarily considers similarity based on weight vectors alone, we argue that gradient information is equally crucial. If two neurons have similar weights but divergent gradients, they are likely to diverge quickly during training without intervention.

Intuitively, higher similarity should induce stronger repulsion. We therefore construct a pairwise repulsion force proportional to both the weight-space cosine similarity and the gradient-space cosine similarity between neurons. Let $\mathbf{w} \in \mathbb{R}^{N \times D}$ denote the weight matrix of a layer with $N$ neurons (each row is the weight vector of one neuron) and $\mathbf{g} \in \mathbb{R}^{N \times D}$ the gradient matrix. We define:
\begin{equation}
    \mathbf{C}^w_{i,j} = \frac{\mathbf{w}_i^\top \mathbf{w}_j}{\Vert \mathbf{w}_i \Vert \Vert \mathbf{w}_j \Vert}, \quad
    \mathbf{C}^g_{i,j} = \frac{\mathbf{g}_i^\top \mathbf{g}_j}{\Vert \mathbf{g}_i \Vert \Vert \mathbf{g}_j \Vert},
    \label{eq: cos_def}
\end{equation}
as the cosine similarity matrices for weights and gradients. The repulsion force for neuron $i$ is then:
\begin{equation}
    (\Delta \mathbf{g})_i = -\sum_{j=1}^{N} \mathbf{C}^w_{i,j} \cdot \mathbf{C}^g_{i,j} \cdot \mathbf{g}_j,
    \label{eq: soft_repulsion}
\end{equation}
which pushes neuron $i$'s update in the opposite direction of the gradients of neurons that are similar to it in both weight and gradient space. This design ensures that the repulsion is strongest when neurons are structurally similar \emph{and} learning similar features, while neurons that are already differentiated or learning distinct patterns receive minimal perturbation. The soft collision update scheme is summarized in Algorithm~\ref{alg:soft}.

\textbf{From hard to soft collisions.} Although Eq.~\ref{eq: soft_repulsion} is written as a repulsion force, it is not an ad hoc construction but a deterministic surrogate motivated by the hard DSMC update of Eq.~\ref{eq: delta_v}. Substituting the center-of-mass gradient $(\mathbf{g}_{\mathrm{cm}})_{i,j}=\tfrac12(\mathbf{g}_i+\mathbf{g}_j)$ into Eq.~\ref{eq: delta_v} decomposes each pairwise collision into a deterministic and a stochastic part,
\begin{equation}
(\Delta \mathbf{g})_{i,j}
= \underbrace{-\tfrac{1}{2}\left(\mathbf{g}_i-\mathbf{g}_j\right)}_{\text{deterministic drift}}
+ \underbrace{\tfrac{1}{2}(\mathbf{g}_r)_{i,j}\,\mathbf{n}_{i,j}}_{\text{stochastic rescattering}} .
\label{eq: hard_decomp}
\end{equation}
Summing the pairwise contributions weighted by the probability that the pair is selected by the DSMC acceptance test of Eq.~\ref{eq: coll_mask} gives the expected update to neuron $i$,
\begin{equation}
\mathbb{E}\!\left[\Delta \mathbf{g}_i\right]
= \sum_{j\neq i} \underbrace{\Pr\!\left[(i,j)\ \text{collide}\right]}_{\text{rate }\lambda_{i,j}}\;
\mathbb{E}_{\mathbf{n}}\!\left[(\Delta \mathbf{g})_{i,j}\right].
\label{eq: expected_update}
\end{equation}
Two approximations reduce this sampling-based update to the differentiable closed form of Eq.~\ref{eq: soft_repulsion}:

\emph{(i) Collision rate as a similarity kernel.} The acceptance rule in Eq.~\ref{eq: coll_mask} fires pair $(i,j)$ with a probability that grows with their gradient interaction $(\mathbf{g}_r)_{i,j}$ and with weight-space proximity through the locality factor $(U_r)_{i,j}=e^{-(\mathbf{w}_r)_{i,j}}$. Replacing the Bernoulli mask by its expected rate $\lambda_{i,j}$ and modelling this rate as separable in a weight-space and a gradient-space proximity term, we adopt the cosine similarities of Eq.~\ref{eq: cos_def}, which are angular rather than metric and hence heuristic, as smooth and scale-free proxies:
\begin{equation}
\lambda_{i,j}\ \approx\ \text{coll}\_{\textrm{coef}}\cdot
\mathbf{C}^{w}_{i,j}\,\mathbf{C}^{g}_{i,j}.
\label{eq: rate_approx}
\end{equation}
Both factors are monotone stand-ins for the DSMC filter: strongly aligned weights correspond to a small $(\mathbf{w}_r)_{i,j}$ and hence a large $(U_r)_{i,j}$, while aligned gradients correspond to a large collision cross-section. Pairs that are already dissimilar in either space receive a vanishing rate and are left essentially untouched.

\emph{(ii) Deterministic repulsive impulse.} We next identify where the separating effect of the hard collision originates. Since $\mathbf{n}_{i,j}$ is drawn uniformly on the sphere, the rescattering term is zero-mean, that is $\mathbb{E}_{\mathbf{n}}[\tfrac12(\mathbf{g}_r)_{i,j}\mathbf{n}_{i,j}]=\mathbf{0}$, and contributes no first-order force; its effect is of second order, corresponding to the anisotropic velocity spread that underlies the entropy-increasing dynamics of the H-theorem interpretation of KO, which will be introduced in the next section. The surviving first moment in Eq.~\ref{eq: expected_update} is therefore the center-of-mass drift $-\tfrac12(\mathbf{g}_i-\mathbf{g}_j)$, which is contractive: it draws the two gradients toward their common mean and would itself promote condensation. The desired separation thus cannot arise from the first-moment (mean-field) limit of the hard update, but originates in its second-order rescattering. The soft collision approximates this second-order effect deterministically, preserving its repulsive sign while replacing the stochastic term by a first-order impulse $-\mathbf{g}_j$ that displaces neuron $i$ away from the gradient direction shared with a similar partner. Substituting Eq.~\ref{eq: rate_approx} and this impulse into Eq.~\ref{eq: expected_update} gives $\Delta \mathbf{g}_i\ \approx\ -\sum_{j=1}^{N}\ \text{coll}\_{\textrm{coef}}\cdot \mathbf{C}^{w}_{i,j}\,\mathbf{C}^{g}_{i,j}\,\mathbf{g}_j$, which recovers the soft repulsion of Eq.~\ref{eq: soft_repulsion} once the coefficient $\text{coll}\_{\textrm{coef}}$ is absorbed into the update step of Algorithm~\ref{alg:soft}. The soft collision is therefore a differentiable surrogate inspired by the hard collision: the stochastic collision selection is replaced by a similarity kernel, and the random rescattering by a deterministic repulsive impulse of matching sign, retaining the PDE-simulation intuition while eliminating the per-step sampling that renders the hard variant expensive.

\begin{algorithm}[!tb]
\caption{Soft Collision-Based Gradient Updates}
\label{alg:soft}
\begin{algorithmic}[1]
\STATE \textbf{Input:} layer weight $\mathbf{w}\in \mathbb{R}^{N\times D}$, layer gradient $\mathbf{g}\in \mathbb{R}^{N\times D}$, hyper-parameter: $\text{coll}\_{\textrm{coef}}$.
\STATE \textbf{Output:} updated layer gradient $\mathbf{g}\in \mathbb{R}^{N\times D}$.
\STATE Compute weight cosine similarity: $\mathbf{C}^w_{i,j} = \frac{\mathbf{w}_i^\top \mathbf{w}_j}{\Vert \mathbf{w}_i \Vert \Vert \mathbf{w}_j \Vert}$ for all $i,j$.
\STATE Compute gradient cosine similarity: $\mathbf{C}^g_{i,j} = \frac{\mathbf{g}_i^\top \mathbf{g}_j}{\Vert \mathbf{g}_i \Vert \Vert \mathbf{g}_j \Vert}$ for all $i,j$.
\STATE Compute repulsion force: $\Delta \mathbf{g}_i = -\sum_{j=1}^{N} \mathbf{C}^w_{i,j} \cdot \mathbf{C}^g_{i,j} \cdot \mathbf{g}_j$ for each neuron $i$.
\STATE Update gradients: $\mathbf{g} = \mathbf{g} + \text{coll}\_{\textrm{coef}} \cdot \Delta \mathbf{g}$;
\STATE \textbf{Return} $\mathbf{g}$;
\end{algorithmic}
\end{algorithm}

\vspace{-5pt}
\subsection{Understanding KO from a Physical Perspective}
To motivate the use of collisions, we provide a physical interpretation. In statistical mechanics, the H-theorem~\citep{boltzmann1970weitere} describes how collisions drive a system toward maximum entropy~\citep{renyi1961measures}, i.e., increasing disorder. As a consequence of the Boltzmann Transport Equation, it states that under molecular chaos, the entropy of an isolated system increases monotonically until equilibrium.

Drawing an analogy to neural networks, we treat neurons as particles, and weight correlation is a measure of system order. High condensation corresponds to low entropy: when neurons collapse to identical values, the system becomes fully ordered. Although neural training does not satisfy the assumptions of thermodynamic systems or the H-theorem, the analogy offers useful intuition: collision-based perturbations can counteract clustering, similar to entropy-increasing collisions in physics. We stress that this interpretation is conceptual rather than formal; rigorous analysis is provided in the next subsection. Table~\ref{tab:conn} summarizes the correspondence between kinetic theory and neural optimization.

\subsection{Understanding KO from a Mathematical Perspective}
\label{section: understand}
We now provide a mathematical analysis of KO's effect on weight correlation. Weight correlation is defined in Eq.~\ref{eq: wc_def} as the average pairwise absolute cosine similarity of weight vectors within a layer. \cite{jin2020doesweightcorrelationaffect} establishes a relationship between weight correlation and generalization within the PAC-Bayesian framework~\citep{10.1145/307400.307435}, where the generalization error can be bounded by the Kullback-Leibler (KL) divergence between posterior and prior distributions of network parameters~\citep{dziugaite2017computingnonvacuousgeneralizationbounds}:

\begin{theorem}[\cite{jin2020doesweightcorrelationaffect}]
For a nontrivial network, a decrease in network weight correlation leads to a reduction in the KL divergence between the posterior and prior distributions of the network parameters, and hence yields a tighter upper bound on the generalization error.
\label{thm:1}
\end{theorem}

Given Thm \ref{thm:1}, we can investigate the generalizability of the method by looking into its influence on the network's weight correlation. We state the result informally here; the full proof with detailed assumptions is provided in Appendix~\ref{appx: proof}.

\begin{theorem}
\label{thm:3}
    Given a sufficiently small learning rate, both collisions reduce the weight correlation of the applied network layer.
\end{theorem}

The theorem confirms that KO achieves its design objective: the collision mechanism provides a structured repulsive force that actively separates correlated neurons. Combined with Theorem~\ref{thm:1}, this result implies that KO's reduction of weight correlation translates to tighter PAC-Bayesian generalization bounds. We note that the analysis makes a small-learning-rate approximation and assumes the training has progressed beyond the initial transient phase; these are standard assumptions in the analysis of learning dynamics~\citep{bottou2018optimization}.

\subsection{Proof of Convergence}
Convergence is a fundamental criterion in evaluating the reliability of optimizers. To ensure that our collision mechanism does not compromise this property, we establish a formal convergence guarantee. The analysis builds on standard smooth optimization arguments; details are provided in Appendix~\ref{appx: proof}.
\begin{theorem}
Let $\mathcal{L}$ be an $L$-smooth loss function bounded below by $\mathcal{L}^*$. Consider parameter updates of the form ${\bm\theta}_{t+1} = \bm\theta_t - \eta (\bm{g}_{\bm\theta_t} + \bm\delta_t)$, where $\bm{g}_{\bm\theta} = \nabla_{\bm\theta} \mathcal{L}$, $\eta > 0$ is the step size, and $\bm\delta_t$ is the collision perturbation.
\begin{itemize}[left=0pt]
    \item If the hard collision, which satisfies $\mathbb{E}[\bm\delta_t] = 0$ and $\mathbb{E}\Vert\bm\delta_t\Vert^2 \le \sigma^2$ for all t, is applied, for any $\varepsilon > 0$, there exists a stepsize $\eta$ such that $\frac{1}{T}\sum_{t=1}^{T}\mathbb{E}[\Vert \bm{g}_{\bm\theta_t} \Vert^2] \le \varepsilon$ after $\mathcal{O}(\frac{\sigma^2}{\varepsilon^2})$ iterations.
    \item If the soft collision, which satisfies $\Vert \bm\delta_t \Vert \le D$ almost surely for all t, is applied, for any $\varepsilon > 0$, there exists a size $\eta$ such that $\frac{1}{T}\sum_{t=1}^{T}\mathbb{E}[\Vert \bm{g}_{\bm\theta_t} \Vert^2] \le \varepsilon$ after $\mathcal{O}(\frac{D^2}{\varepsilon^2})$ iterations.
\end{itemize}
\label{thm:convergence}
\end{theorem}
This theorem is proved with a traditional gradient descent optimizer as the base optimizer. Due to the boundedness of the collision term incorporated, we state that the convergence analysis on the momentum, Adam-like, or other types of gradient descent schemes can be directly applied to those with the collision introduced. It will only incur a constant level change in the final result. We note that this result establishes that KO does not harm convergence rather than claiming a faster rate; the benefit of KO lies in the improved generalization characterized by Theorem~\ref{thm:3} and Theorem~\ref{thm:1}.

\vspace{-5pt}
\section{Experiments}
\label{sec:experiments}
\subsection{Condensation Effects Study}
In this section, we validate the anti-condensation effects of the proposed methods. We follow the experiment settings of \cite{zhou2022understandingcondensationneuralnetworks}. We use the cosine similarity of neuron weights to evaluate neuronal similarity, with the Adam optimizer~\citep{kingma2014adam} as the base optimizer. We visualize the cosine similarity matrix after clustering-based reordering: starting from neuron 0, we iteratively group neurons that are strongly similar (cosine similarity $> 0.6$), reorder the matrix accordingly, and display the result. Large dense diagonal blocks indicate strong condensation into clusters; a more fragmented pattern indicates lower correlation. A detailed description of this visualization paradigm is in Appendix~\ref{appx:condense}.

\begin{figure*}[tb!]
    \centering
    \subfigure[Tanh]{
        \begin{minipage}[b]{0.235\linewidth}
    \includegraphics[width=1\linewidth]{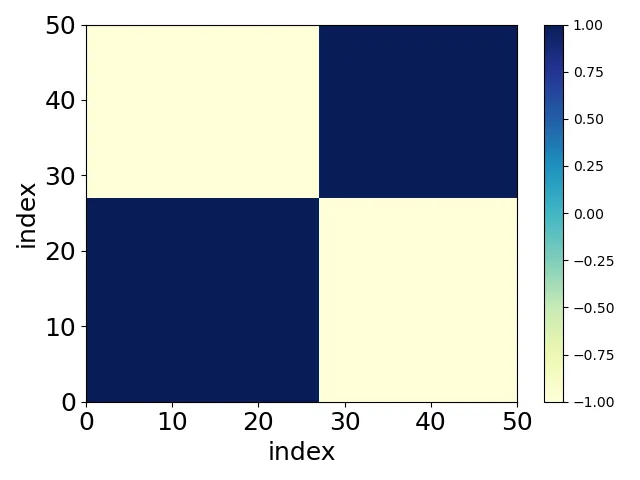}\vspace{4pt}
    \includegraphics[width=1\linewidth]{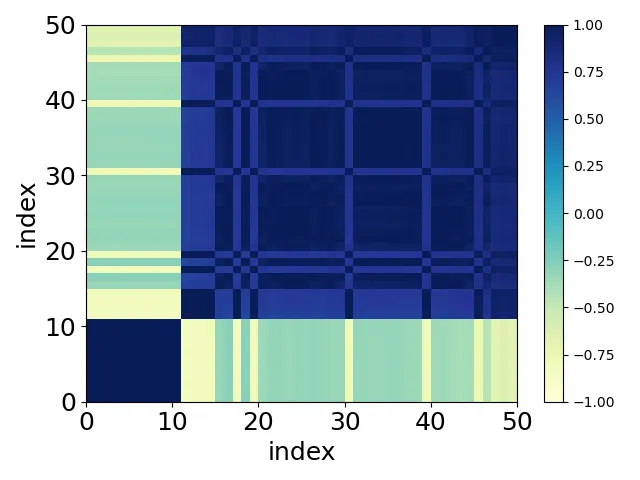}
    \includegraphics[width=1\linewidth]{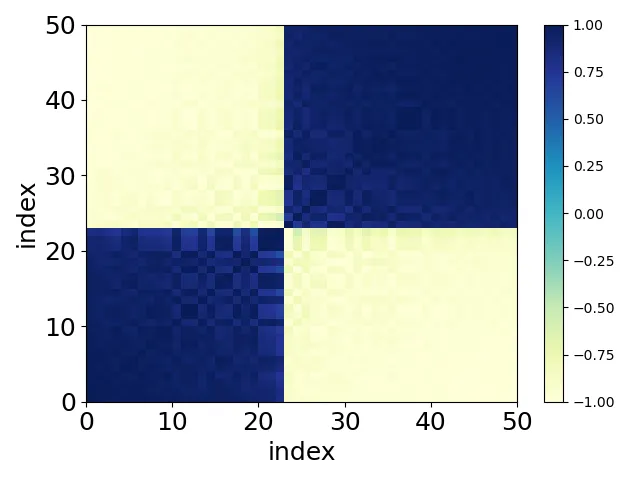}
    \end{minipage}}
    \subfigure[xTanh]{
    \begin{minipage}[b]{0.235\linewidth}
    \includegraphics[width=1\linewidth]{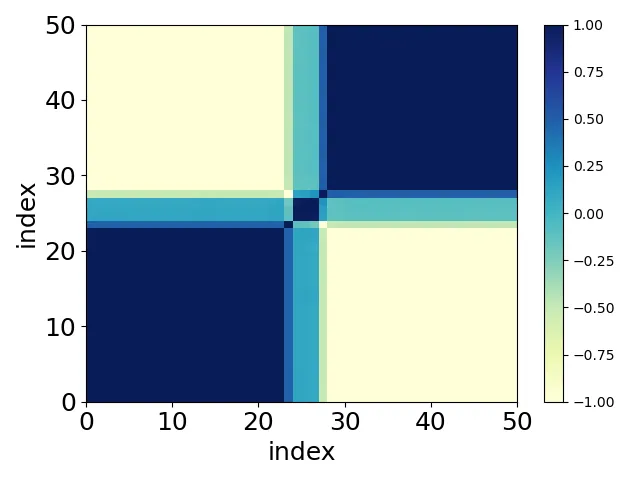}\vspace{4pt}
    \includegraphics[width=1\linewidth]{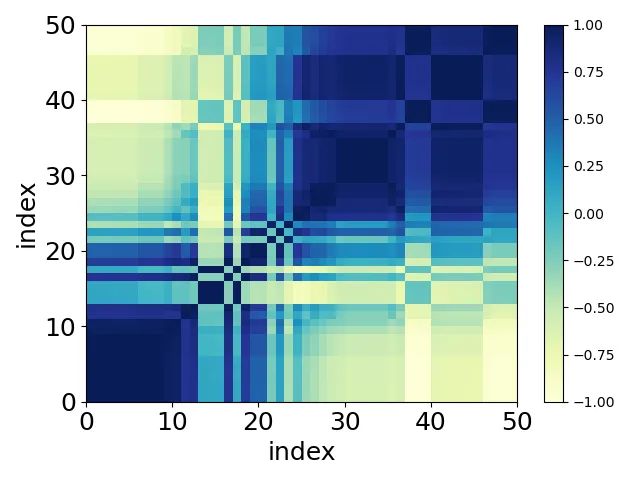}
    \includegraphics[width=1\linewidth]{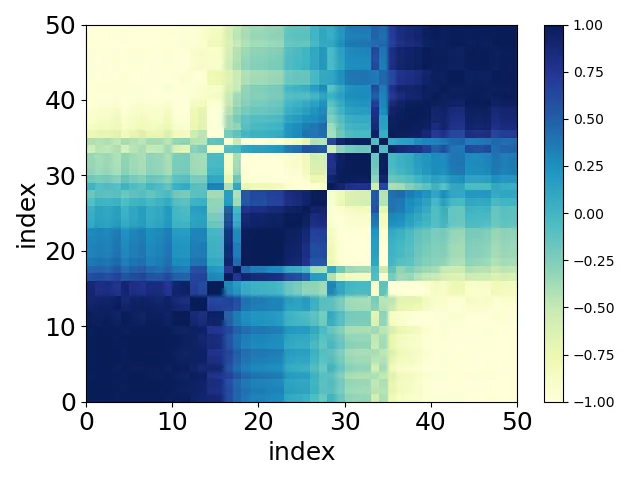}
        \end{minipage}}
\subfigure[Sigmoid]{
    \begin{minipage}[b]{0.235\linewidth}
    \includegraphics[width=1\linewidth]{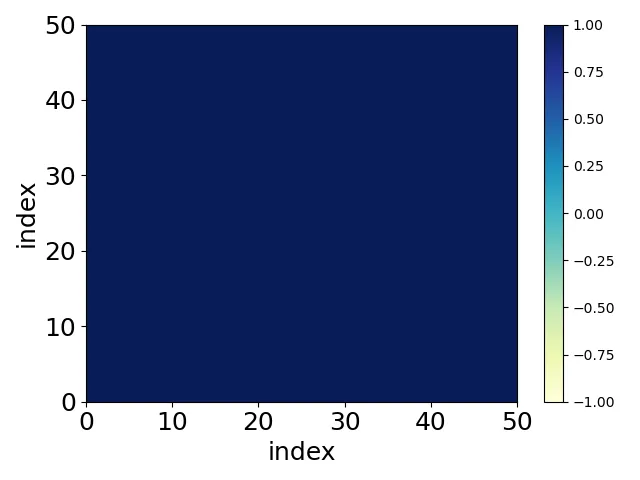}\vspace{4pt}
    \includegraphics[width=1\linewidth]{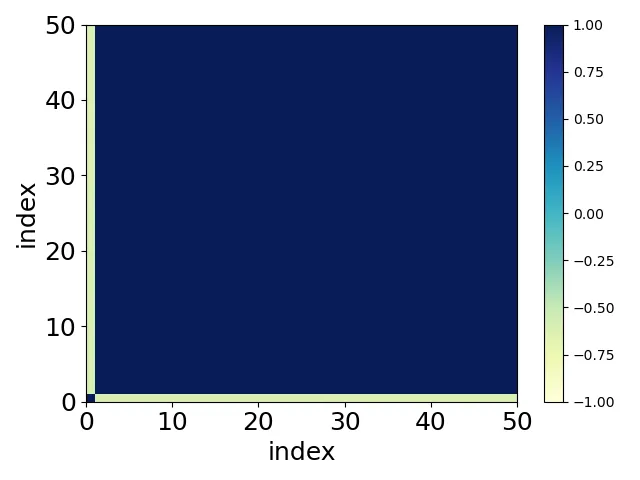}
    \includegraphics[width=1\linewidth]{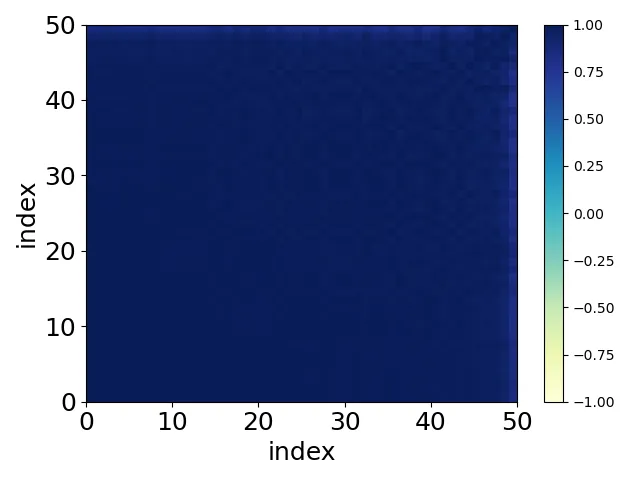}
\end{minipage}}
\subfigure[Softplus]{
    \begin{minipage}[b]{0.235\linewidth}
    \includegraphics[width=1\linewidth]{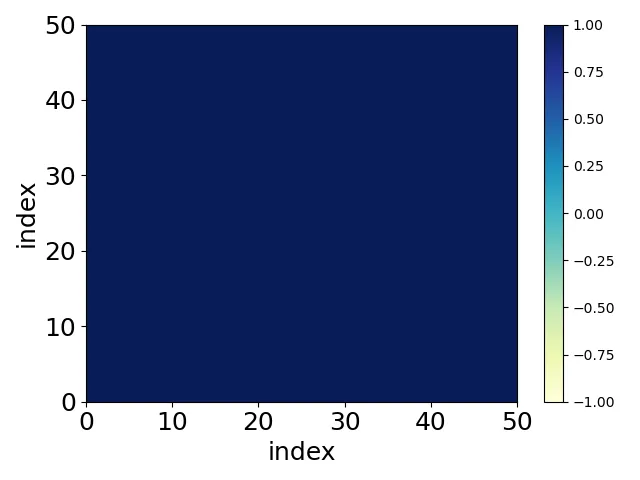}\vspace{4pt}
    \includegraphics[width=1\linewidth]{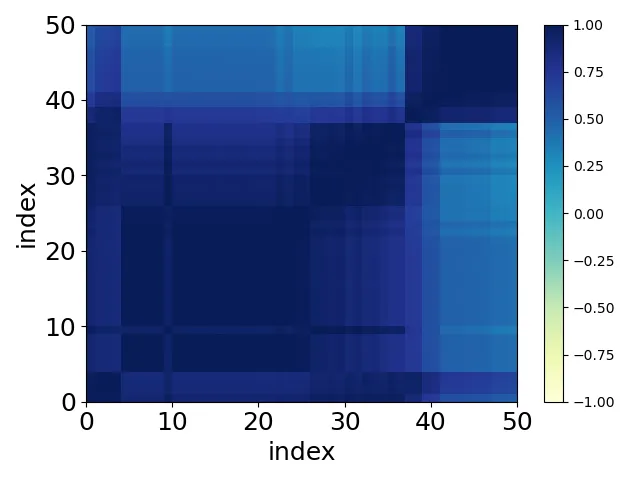}
    \includegraphics[width=1\linewidth]{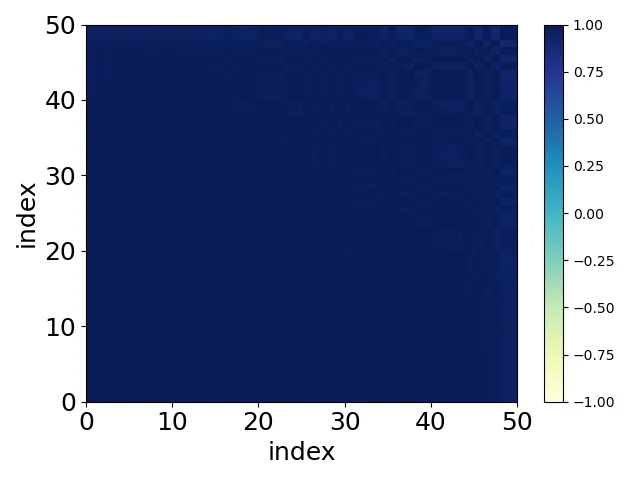}
\end{minipage}}
\vspace{-5pt}
\caption{Condensation of two-layer NNs. The color indicates the cosine similarity of two hidden neurons’ input weights at epoch 100, whose indices are indicated by the abscissa and the ordinate, respectively. The activation functions are indicated by the sub-captions. The first row shows the model weights trained with the original Adam optimizer. The second and third rows depict the results with the Soft Collision and Hard Collision. A more detailed description is in Appendix \ref{appx:condense}.}
\label{fig:2layer}
\end{figure*}

\textbf{Experiments on a Multidimensional Synthetic Dataset} Firstly, we perform simple experiments on a multidimensional synthetic dataset. As in \cite{zhou2022understandingcondensationneuralnetworks}, we utilize a 2-layer fully-connected network with a hidden dimension 5-50-1 to fit 80 instances sampled from a 5-dimensional function $\sum_{k=1}^5 3.5 \sin (5x_k+1)$, where each $x_k$ is sampled uniformly from $[-4, 2]$. All parameters are initialized from a Gaussian distribution $\mathcal{N}(0, 0.005^2)$. 

The result of the experiment is shown in Fig. \ref{fig:2layer}. It can be seen that both the proposed hard collision and soft collision hinder the condensation of the weights. This result is in line with the mathematical proof and the physical interpretation. Further condensation experiments are shown in Appendix \ref{appx:condense}. We could draw a simple conclusion in this section that the collision optimizers empirically halt the condensation in the network weights. 

\vspace{-5pt}
\subsection{Experiments on Image Datasets}
\textbf{Experiments on CIFAR-10 and CIFAR-100.}
Firstly, we perform experiments on two image datasets, CIFAR-10 and CIFAR-100~\citep{krizhevsky2009learning}. ResNets of various sizes are chosen as the model backbone~\citep{he2015deepresiduallearningimage}. A ResNet model can be separated into a convolutional encoder and an FC-layer-based classifier. Since both hard and soft collisions are now designed explicitly for fully connected layers, we apply our proposed collision mechanism to the classifier layer to avoid condensation in that layer. Besides making a comparison with the vanilla model, we further compare with the penalty method that controls weight correlation in \cite{jin2020doesweightcorrelationaffect}.

The training process incorporates the SGD optimizer with a weight decay of $5 \times 10^{-4}$, a batch size of 128, a momentum of 0.9, and an initial learning rate of 0.1. A cosine annealing learning rate scheduler is further adopted to better control the learning process. All models are trained for 200 epochs. Each image is padded by 4 pixels on each side, followed by random cropping of a 32 × 32 section from the padded image or its horizontal flip, as in the original paper. The prediction accuracy is chosen as the evaluation metric.

\begin{table}[]
    \centering
    \caption{Accuracy on CIFAR datasets. S.C. stands for Soft Collision, H.C. for Hard Collision, and WCreg for the weight correlation regularization.}
    \vspace{-5pt}
    \begin{tabular}{c|c|c}
        \toprule
         Model & CIFAR10 & CIFAR100\\
         \midrule
         ResNet18+SGD & $95.07\% \pm 0.18\%$ & $78.69\% \pm 0.38\%$\\
         ResNet18+S.C. & \textbf{$\bm{95.74\% \pm 0.05\%}$}& \textbf{$\bm{79.08\% \pm 0.12\%}$}\\
         ResNet18+H.C. & $95.58\% \pm 0.07\%$ & $78.83\% \pm 0.09\%$\\
         ResNet18+WCreg & $95.51\% \pm 0.23\%$ & $78.77\% \pm 0.28\%$ \\
         \midrule
         ResNet34+SGD & $95.14\% \pm 0.37\%$ & $79.42\% \pm 0.45\%$\\
         ResNet34+S.C. & \textbf{\bm{$95.76\% \pm 0.08\%$}}& \textbf{$\bm{79.88\% \pm 0.25\%}$}\\
         ResNet34+H.C. & $95.56\% \pm 0.14\%$ & $79.57\% \pm 0.32\%$\\
         ResNet34+WCreg & $95.68\% \pm 0.29\%$ & $79.66\% \pm 0.11\%$\\
         \midrule
         ResNet50+SGD & $95.37\% \pm 0.38\%$ & $78.59\% \pm 0.56\%$\\
         ResNet50+S.C. & \textbf{$\bm{95.83\% \pm 0.16\%}$} & \textbf{$\bm{79.27\% \pm 0.23\%}$}\\
         ResNet50+H.C. & $95.57\% \pm 0.05\%$ & $78.98\% \pm 0.18\%$\\
         ResNet50+WCreg &  $95.42\% \pm 0.24\%$ & $79.01\% \pm 0.20\%$\\
         \bottomrule
    \end{tabular}
    \label{tab:cifar}
    \vspace{-10pt}
\end{table}

Table \ref{tab:cifar} presents results on CIFAR datasets. The introduction of the collision mechanism boosts the models' performance. Comparatively, the soft collision improves the test accuracy to a larger extent compared to the hard version. The soft collision approximately increases $0.5\%$ on the test accuracy, while the hard collision raises the accuracy by roughly $0.3\%$. Furthermore, the incorporation of collision enables shallower models to win over deeper models without collision. On CIFAR-10, ResNet18+Soft Collision predicts better than a vanilla ResNet50. This means that the introduction of this simple optimizer improvement is more effective than merely increasing network depth.

This improvement is notable since the collision mechanism basically introduces no extra time to the training and doesn't need to modify the model structure, the dataset (no extra augmentation is required), or the training scheme. Specifically, it takes 3.68h to train a ResNet50 with the vanilla SGD optimizer and only 3.69h to train a ResNet50 with SGD and Soft Collision. The increase is negligible compared to the duration of the entire training. 

\begin{figure*}[!tb]
    \centering
    \subfigure[ResNet18]{
        \begin{minipage}[b]{0.32\linewidth}
    \includegraphics[width=1\linewidth]{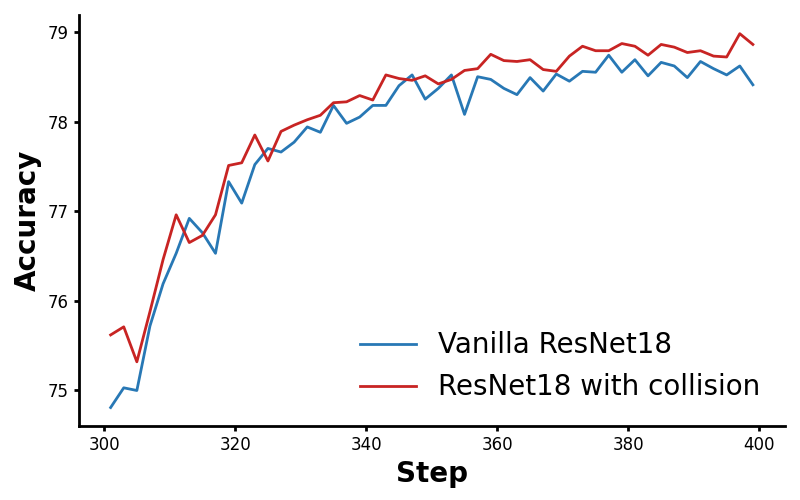}\vspace{4pt}
    \includegraphics[width=1\linewidth]{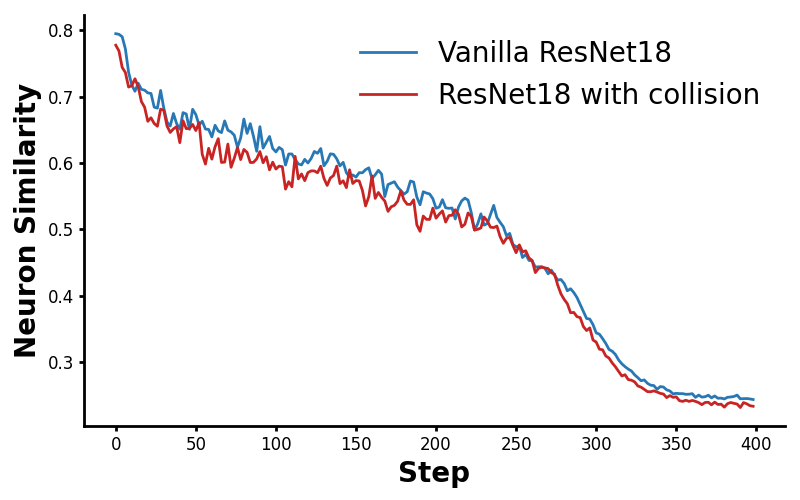}
    \end{minipage}}
    \subfigure[ResNet34]{
    \begin{minipage}[b]{0.32\linewidth}
    \includegraphics[width=1\linewidth]{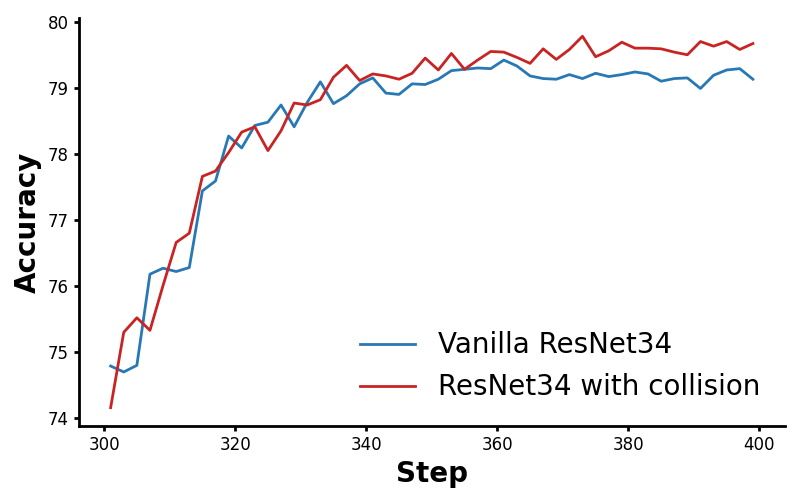}\vspace{4pt}
    \includegraphics[width=1\linewidth]{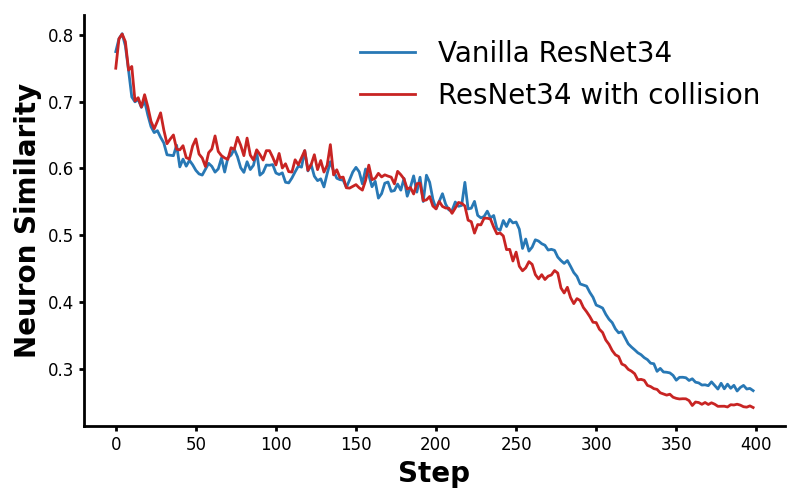}
        \end{minipage}}
\subfigure[ResNet50]{
    \begin{minipage}[b]{0.32\linewidth}
    \includegraphics[width=1\linewidth]{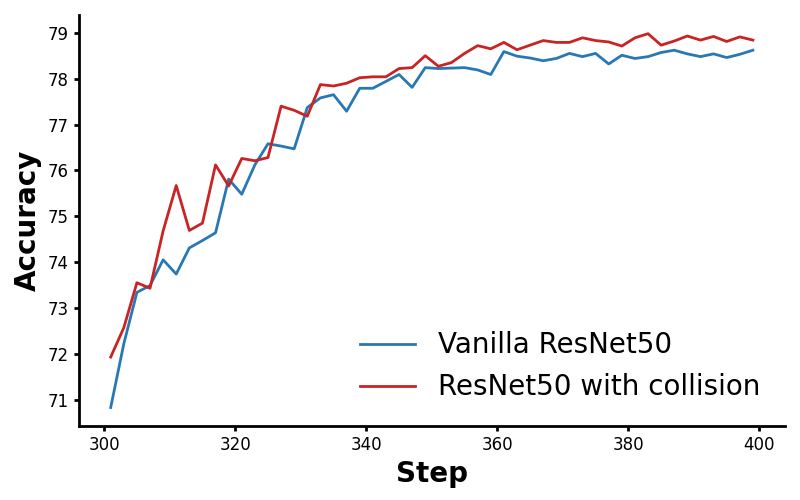}\vspace{4pt}
    \includegraphics[width=1\linewidth]{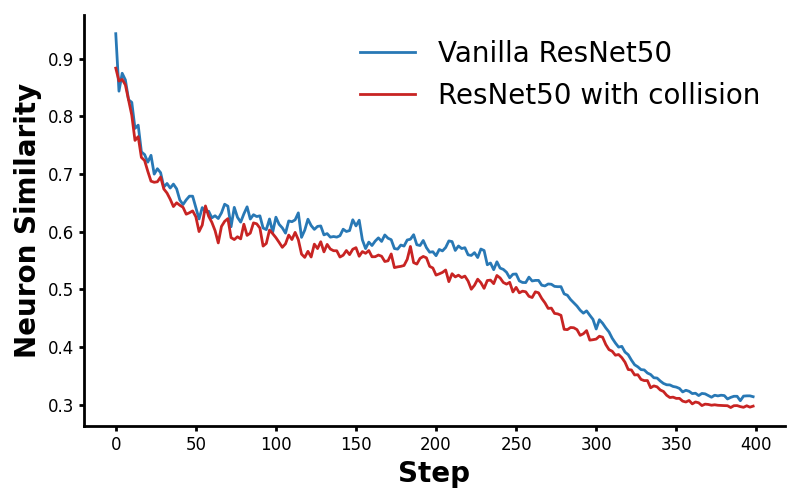}
\end{minipage}}
\caption{Accuracy and neuron similarity of three different models on CIFAR-100. The upper row shows the test accuracy, while the lower row depicts the neuron similarity changes during training.}
\label{fig:acc}
\end{figure*}

To further show our methods' influence on the models, we plot the validation accuracy and neuron similarity changes during the training phase in Fig. \ref{fig:acc}. The neuron similarity is defined in Eq. \ref{eq: wc_def}.


\begin{table*}[tb]
    \centering
    \vspace{-5pt}
    \caption{Comparison of model performance on various LLM benchmarks.}
    \vspace{-5pt}
    \label{tab:experiment_results}
    \begin{tabular}{lccccccc}
        \toprule
        Model $\setminus$ Acc $\uparrow$ & PIQA & ARC\_C & ARC\_E & HellaSwag & WinoGrande & OBQA & Avg \\
        \midrule
        Qwen2-0.5B+AdamW & 0.575 & 0.227 & 0.371 & 0.273 & 0.498 & 0.246 & 0.365 \\
        Qwen2-0.5B+S.C. & \textbf{0.582} & \textbf{0.241} & \textbf{0.385} & \textbf{0.276} & \textbf{0.522} & \textbf{0.268} & \textbf{0.379} \\
        \midrule
        Qwen2-1.5B+AdamW & 0.602 & 0.236 & 0.439 & 0.310 & 0.495 & 0.256 & 0.390 \\
        Qwen2-1.5B+S.C. & \textbf{0.615} & \textbf{0.258} & \textbf{0.450} & \textbf{0.313} & \textbf{0.535} & \textbf{0.266} & \textbf{0.406} \\
        \bottomrule
    \end{tabular}
    \vspace{-5pt}
\end{table*}

Fig.~\ref{fig:acc} shows that our method effectively reduces weight similarity, as indicated in Section \ref{section: understand}. The collision mechanism mitigates condensation and maintains lower inter-neuron similarity throughout training, even on this standard benchmark. Notably, performance gains over the vanilla ResNet are most pronounced in the final training epochs, likely due to the interaction between cosine annealing and the collision mechanism.

\begin{table}
    \centering
    \caption{Experimental result on ImageNet-1K.}
    \vspace{-10pt}
    \begin{tabular}{c|c|c}
    \toprule
        Model & T-1 ACC.(\%) & T-5 ACC.(\%)\\
        \midrule
        ResNet50+Lamb & 79.70\% & 94.53\%\\
        ResNet50+S.C. & \textbf{80.23\%} & \textbf{94.79\%}\\
        ResNet50+H.C. & 79.98\% & 94.61\%\\
        \midrule
        ConvNext\_Tiny+AdamW & 82.10\% & 96.03\%\\
        ConvNext\_Tiny+S.C. & \textbf{82.34\%} & \textbf{96.07\%}\\
        ConvNext\_Tiny+H.C. & 82.17\% & 96.03\%\\
        \midrule
        ConvNext\_Small+AdamW & 82.83\% & 96.29\%\\
        ConvNext\_Small+S.C. & \textbf{83.12\%} & \textbf{96.33\%} \\
        ConvNext\_Small+H.C. & 82.91\% & 96.29\%\\
        \bottomrule
    \end{tabular}
    \label{tab:imagenet}
    \vspace{-10pt}
\end{table}

\textbf{Experiments on ImageNet-1K}.
In this section, we evaluate our model on a larger image classification dataset, ImageNet-1K~\citep{deng2009imagenet}. ImageNet-1K includes 1.28 million training images and 50,000 validation images across 1,000 categories. Following \citep{wightman2021resnetstrikesbackimproved}, we adopt a series of data augmentation and regularization strategies, including but not limited to RandAugment~\citep{cubuk2019randaugmentpracticalautomateddata}, Mixup alpha~\citep{zhang2018mixupempiricalriskminimization}, and CutMix alpha~\citep{yun2019cutmixregularizationstrategytrain}. The model training script is based on timm~\citep{rw2019timm}. 

We adopt ResNet and ConvNext~\citep{Liu_2022_CVPR} as the training backbone. The ResNet 50 training hyperparameters are set as in \cite{wightman2021resnetstrikesbackimproved}. We follow the ConvNext hyperparameters as in \cite{liu2024dendritic}. For both models, we still only apply collision to the last classifier layer. This is the cleanest method for applying this collision mechanism targeted at fully connected layers. Following the standard evaluation protocol commonly adopted for ImageNet-scale experiments, we report the Top-1 and Top-5 accuracy from a single training run due to the substantial computational cost of repeated full-scale training.

Table~\ref{tab:imagenet} reports ImageNet-1K results. Introducing collision—both soft and hard variants—consistently improves Top-1 and Top-5 accuracy. Although the gains are modest (0.2–0.5\% Top-1), they are meaningful given the maturity of the benchmark and strong training recipes (RandAugment, Mixup, CutMix). Improvements are observed in both ResNet and ConvNext architectures, indicating good generalization. Due to the training costs, results are based on single runs following standard practice~\citep{wightman2021resnetstrikesbackimproved}; multi-seed CIFAR experiments confirm the statistical significance of the gains.

\vspace{-10pt}
\subsection{Experiments on Text Datasets}

\textbf{Experiments on IMDB and Snips.} In this section, we attempt to show that our method works not only on image tasks but also on text-related tasks. We choose two simple classification datasets, IMDB~\citep{maas-EtAl:2011:ACL-HLT2011} and Snips~\citep{coucke2018snipsvoiceplatformembedded}. IMDB comprises 25,000 highly polar movie reviews for training, and 25,000 for testing. Snips is a dataset of over 16,000 crowdsourced queries distributed among seven user intents for intent classification.

As with many text-based tasks, we train a sentiment classifier with a pretrained backbone. We mainly utilize BERT~\citep{devlin2019bertpretrainingdeepbidirectional} and DistilBert~\citep{sanh2020distilbertdistilledversionbert} from the Transformers library by Hugging Face~\citep{wolf2020huggingfacestransformersstateoftheartnatural}. The classification model consists of a Bert Model transformer with a sequence classification head on top (a fully-connected layer). The collision is applied only to the classification head. Each classification model is trained for 10 epochs using an AdamW optimizer~\citep{loshchilov2019decoupledweightdecayregularization}.

The results are shown in Table \ref{tab:txt}. On both of the datasets, collision improves the overall model performance. This result further proves that the collision mechanism works universally. 

\begin{table}
    \centering
    \vspace{-5pt}
    \caption{Accuracy on IMDB and Snips.}
    \vspace{-5pt}
    {
    \begin{tabular}{c|c|c}
    \toprule
        Pretrained Model & IMDB & Snips \\
        \midrule
        bert-base-cased+AdamW & 92.52\% & 98.14\%\\
        bert-base-cased+S.C. & \textbf{93.80\%} & \textbf{98.43\%}\\
        bert-base-cased+H.C. & 93.42\% & \textbf{98.43\%}\\
        \midrule
        bert-base-uncased+AdamW& 93.59\% & 97.71\%\\
        bert-base-uncased+S.C.& \textbf{94.20\%} & \textbf{97.89\%}\\
        bert-base-uncased+H.C.& 93.76\%  & 97.73\%\\
        \midrule
        distilbert-base-uncased+AdamW& 93.07\% & 97.71\%\\
        distilbert-base-uncased+S.C.& \textbf{93.33\%} & \textbf{98.43\%}\\
        distilbert-base-uncased+H.C.& 93.24\% & 98.11\%\\
        \bottomrule
    \end{tabular}
    }
    \label{tab:txt}
\vspace{-15pt}
\end{table}

\textbf{LLM Experiment.}
In this section, we conducted pre-training experiments on Large Language Models (LLMs). Specifically, we trained Qwen2-0.5B (50B tokens) and Qwen2-1.5B (25B tokens) on the FineWeb-Edu dataset with a context length of 4,096. The base optimizer is AdamW~\citep{loshchilov2017decoupled} ($\beta_1=0.9$, $\beta_2=0.95$, weight decay $0.1$), and our proposed soft collision method (S.C.) is applied to the final transformer layer's output projection. The learning rate was linearly warmed up for 3,000 updates to $10^{-4}$ and cosine-decayed to $10^{-5}$ over 200,000 updates. The effective batch contained 120 sequences (491,520 tokens) per update. We evaluate on a diverse set of benchmarks: knowledge-intensive tasks (ARC-C, ARC-E) and commonsense reasoning (HellaSwag, WinoGrande, OBQA, PIQA).

As shown in Table~\ref{tab:experiment_results}, our method consistently outperforms the baseline across all tasks, achieving an average absolute improvement of $>1.4\%$. This gain is approximately equivalent to what would be obtained by training the baseline for roughly 10\% more tokens, while KO incurs only $\sim 3.3\%$ additional wall-clock time. This favorable cost--benefit trade-off demonstrates KO's scalability to large-scale training regimes and its practical value beyond vision tasks. Results are reported from single training runs due to the computational cost of LLM pretraining; the consistent pattern across six diverse benchmarks and two model scales supports the robustness of the gains.

\section{Conclusion and Limitations}
We presented KO, a novel neural optimizer gadget that reinterprets parameter training through the lens of kinetic theory. By modeling parameter updates as stochastic particle collisions, KO introduces a physics-driven mechanism that mitigates weight condensation, a phenomenon closely linked to degraded generalization. Our theoretical analysis establishes that KO provably reduces weight correlation while preserving convergence guarantees, and extensive experiments across vision and language tasks demonstrate consistent accuracy improvements over competitive baselines with negligible computational overhead (for the soft collision variant).

\textbf{Limitations.} KO incurs additional computational overhead due to the simulation of collision dynamics. Moreover, its extension to broader physical formulations and its interaction with modern neural architectures remain open directions.


\newpage
\bibliographystyle{ACM-Reference-Format}
\bibliography{sample-base}

\newpage
\appendix
\onecolumn
\section{Proof}
\label{appx: proof}
\begin{proof}[\textbf{Proof of Thm. \ref{thm:3}}]
Analyzing the changes in the weight correlation of the network layer is equivalent to analyzing the cosine similarity matrix of the layer's weights. We term the weight matrix as $\mathbf{w}$ and its i-th and j-th columns as $\mathbf{w}_i$ and $\mathbf{w}_j$ respectively. The correlation of the weight matrix is defined as in Eq. \ref{eq: wc_def}. Their corresponding gradients are $\mathbf{g}_i$ and $\mathbf{g}_j$. After a step of gradient descent, we have:
\begin{equation}
    \mathbf{w}_i'=\mathbf{w}_i-\eta (\mathbf{g}_i + \Delta_i), \mathbf{w}_j'=\mathbf{w}_j - \eta(\mathbf{g}_j + \Delta_j)
\end{equation}
where $\Delta_i$ and $\Delta_j$ are the collision terms we apply to these two neurons, and $\eta$ is the learning rate. We will look further into the cosine similarity changes after the gradient descent to prove our methods' capacity. The weight cosine similarity can be calculated as:
\begin{equation}
    C_{i, j} = \cos(\mathbf{w}_i, \mathbf{w}_j)=\frac{\mathbf{w}_i^\top \mathbf{w}_j}{\Vert \mathbf{w}_i \Vert \Vert \mathbf{w}_j \Vert}, C_{i,j}'= \frac{(\mathbf{w}_i-\eta (\mathbf{g}_i + \Delta_i))^\top(\mathbf{w}_j - \eta(\mathbf{g}_j + \Delta_j))}{\Vert \mathbf{w}_i-\eta (\mathbf{g}_i + \Delta_i) \Vert \Vert \mathbf{w}_j - \eta(\mathbf{g}_j + \Delta_j) \Vert}
\end{equation}
We assume that the learning rate $\eta$ is a small number. Then we can use Taylor's expansion to get the following results:
\begin{equation}
\begin{aligned}
    C_{i,j}'&\approx \frac{(\mathbf{w}_i^\top \mathbf{w}_j - \eta \mathbf{w}_i^\top (\mathbf{g}_j + \Delta_j) - \eta (\mathbf{g}_i + \Delta_i)^\top \mathbf{w}_j)(1 + \eta \frac{\mathbf{w}_i^\top}{\Vert \mathbf{w}_i \Vert ^2} (\mathbf{g}_i + \Delta_i) )(1 + \eta \frac{\mathbf{w}_j^\top}{\Vert \mathbf{w}_j \Vert ^2} (\mathbf{g}_j + \Delta_j))}{\Vert \mathbf{w}_i \Vert \Vert \mathbf{w}_j \Vert}\\
    &=\frac{\mathbf{w}_i^\top \mathbf{w}_j (1 + \eta \frac{\mathbf{w}_j^\top}{\Vert \mathbf{w}_j \Vert ^2} (\mathbf{g}_j + \Delta_j) + \eta \frac{\mathbf{w}_i^\top}{\Vert \mathbf{w}_i \Vert ^2} (\mathbf{g}_i + \Delta_i)) - \eta \mathbf{w}_i^\top (\mathbf{g}_j + \Delta_j) - \eta (\mathbf{g}_i + \Delta_i)^\top \mathbf{w}_j}{\Vert \mathbf{w}_i \Vert \Vert \mathbf{w}_j \Vert} \\
    & = C_{i,j} + \eta \frac{\mathbf{w}_i^\top \mathbf{w}_j (\frac{\mathbf{w}_j^\top}{\Vert \mathbf{w}_j \Vert ^2} (\mathbf{g}_j + \Delta_j) + \frac{\mathbf{w}_i^\top}{\Vert \mathbf{w}_i \Vert ^2} (\mathbf{g}_i + \Delta_i)) - \mathbf{w}_i^\top (\mathbf{g}_j + \Delta_j) - (\mathbf{g}_i + \Delta_i)^\top \mathbf{w}_j}{\Vert \mathbf{w}_i \Vert \Vert \mathbf{w}_j \Vert} \\
    & = C_{i,j} + \eta \frac{\mathbf{w}_i^\top \mathbf{w}_j (\frac{\mathbf{w}_j^\top}{\Vert \mathbf{w}_j \Vert} \mathbf{g}_j + \frac{\mathbf{w}_i^\top}{\Vert \mathbf{w}_i \Vert} \mathbf{g}_i) - \mathbf{w}_i^\top \mathbf{g}_j - \mathbf{g}_i^\top \mathbf{w}_j}{\Vert \mathbf{w}_i \Vert \Vert \mathbf{w}_j \Vert} \\ & + \eta \frac{\mathbf{w}_i^\top \mathbf{w}_j (\frac{\mathbf{w}_j^\top}{\Vert \mathbf{w}_j \Vert ^2} \Delta_j + \frac{\mathbf{w}_i^\top}{\Vert \mathbf{w}_i \Vert ^2}\Delta_i) - \mathbf{w}_i^\top \Delta_j - \Delta_i^\top \mathbf{w}_j}{\Vert \mathbf{w}_i \Vert \Vert \mathbf{w}_j \Vert}
\end{aligned}
\label{eq:c}
\end{equation}
Next, we closely examine the last term in the above equation. The first two terms in Eq. \ref{eq:c} represent the original cosine similarity after the updates. We only need to prove that our collision can hinder condensation on top of the original pipelines. We term $C_{i, j}'-C_{i,j}$ as $\delta C_{i,j}$.

\textbf{Soft Collision} We first examine the influence of the soft collision. We make a simple assumption that a collision only happens between $\mathbf{w}_i$ and $\mathbf{w}_j$. Then we can simplify the collision term to the following format:
\begin{equation}
    \Delta_j = \alpha \mathbf{g}_i, \Delta_i = \alpha \mathbf{g}_j
\end{equation}
where $\alpha$ is related to the similarity between neurons. We insert this into Eq. \ref{eq:c}.
\begin{equation}
    \begin{aligned}
        \delta C_{i,j} &= \mathbf{w}_i^\top \mathbf{w}_j (\frac{\mathbf{w}_j^\top}{\Vert \mathbf{w}_j \Vert ^2} \alpha \mathbf{g}_i + \frac{\mathbf{w}_i^\top}{\Vert \mathbf{w}_i \Vert ^2}\alpha \mathbf{g}_j) - \mathbf{w}_i^\top \alpha \mathbf{g}_i - \alpha \mathbf{g}_j^\top \mathbf{w}_j \\ 
        & = \alpha\left[ (\cos(\mathbf{w}_i, \mathbf{w}_j) \frac{\Vert \mathbf{w}_i \Vert}{\Vert \mathbf{w}_j \Vert}\mathbf{w}_j - \mathbf{w}_i)^\top \mathbf{g}_i + (\cos(\mathbf{w}_i, \mathbf{w}_j) \frac{\Vert \mathbf{w}_j \Vert}{\Vert \mathbf{w}_i \Vert}\mathbf{w}_i - \mathbf{w}_j)^\top \mathbf{g}_j\right]
    \end{aligned}
    \label{eq:ap2}
\end{equation}
For simplicity, we focus on the training phase when it becomes stable; hence, we have $\mathbf{w}_i^\top \mathbf{g}_i < 0$ and $\mathbf{w}_j^\top \mathbf{g}_j < 0$. This is because when training is stabilized, $\mathbf{g}_i$ won't push $\mathbf{w}_i$ too harshly. We find that we can simplify the analysis of the first term in Eq. \ref{eq:ap2} to determine the sign of $\cos(\mathbf{w}_i, \mathbf{w}_j)\cos(\mathbf{w}_j, \mathbf{g}_i) - \cos(\mathbf{w}_i, \mathbf{g}_i)$ as $\Vert\frac{\Vert \mathbf{w}_i \Vert}{\Vert \mathbf{w}_j \Vert}\mathbf{w}_j\Vert=\Vert \mathbf{w}_i \Vert$. We make an assumption that $\theta(\mathbf{w}_i, \mathbf{g}_i) < \theta(\mathbf{w}_j, \mathbf{g}_i)$. The justification of this assumption is that $\mathbf{g}_i$ is the gradient of $\mathbf{w}_i$ and hence they are more correlated. All the analysis is based on the fact that we only analyze the dynamics in the stable training phase. In such a scenario, the angle between the weight vector $\mathbf{w}_i$ and its corresponding gradient $g_i$ is expected to be small, indicating convergence toward a local minimum. Then $\cos(\mathbf{w}_i, \mathbf{w}_j)\cos(\mathbf{w}_j, \mathbf{g}_i) - \cos(\mathbf{w}_i, \mathbf{g}_i) < 0$. Since $\alpha$ is negatively correlated with neuron similarity, then $\delta C_{i,j}$ pushes $C_{i,j}$ in the opposite direction of $C_{i,j}$ and hence lowers the overall cosine similarity. Conclusively, our soft collision helps to lower the overall weight correlation.

\textbf{Hard Collision} Next, we analyze how the hard collision functions. In the process of backpropagation, the negative gradient serves as the speed, and the weight matrix serves as the position. The hard collision procedure is detailed in the main paper. We still focus on the single collision between $\mathbf{w}_i$ and $\mathbf{w}_j$. The collision term is hence:
\begin{equation}
    \Delta_i = \frac{\mathbf{g}_j - \mathbf{g}_i}{2}+\frac{1}{2}\Vert \mathbf{g}_i - \mathbf{g}_j\Vert \mathbf{u}, \quad \Delta_j = \frac{\mathbf{g}_i - \mathbf{g}_j}{2}-\frac{1}{2}\Vert \mathbf{g}_i - \mathbf{g}_j\Vert \mathbf{u}
\end{equation}
where $\mathbf{u}$ is a random vector with no correlation with $\mathbf{w}$. We combine this with Eq. \ref{eq:c}:
\begin{equation}
    \begin{aligned}
        \delta C_{i,j} = (\cos(\mathbf{w}_i, \mathbf{w}_j) \frac{\Vert \mathbf{w}_i \Vert}{\Vert \mathbf{w}_j \Vert}\mathbf{w}_j - \mathbf{w}_i -\cos(\mathbf{w}_i, \mathbf{w}_j) \frac{\Vert \mathbf{w}_j \Vert}{\Vert \mathbf{w}_i \Vert}\mathbf{w}_i + \mathbf{w}_j)^\top (\frac{\mathbf{g}_i - \mathbf{g}_j}{2}-\frac{1}{2}\Vert \mathbf{g}_i - \mathbf{g}_j\Vert \mathbf{u})
    \end{aligned}
\end{equation}
Since $\mathbf{u}$ is a random vector, on average $-\frac{1}{2}\Vert \mathbf{g}_i - \mathbf{g}_j\Vert u$ will not bring any influence to $\delta C_{i,j}$. We only need to evaluate the terms that include $\frac{\mathbf{g}_j - \mathbf{g}_i}{2}$. Another important condition is that the collision only happens when $\mathbf{w}_i$ and $\mathbf{w}_j$ are close enough and have a large relative speed. This means that $(\mathbf{w}_j - \mathbf{w}_i)^\top (\mathbf{g}_j - \mathbf{g}_i) > 0$ and $\cos (\mathbf{w}_i, \mathbf{w}_j) > 0$. Since the original correlation is positive, we want a $\delta C_{i,j} < 0$ to decrease their weight similarity. Given the above assumption, we can turn to analyze the sign of $(\mathbf{w}_j - \mathbf{w}_i)^\top(\mathbf{g}_i-\mathbf{g}_j)$. Given that $(\mathbf{w}_j - \mathbf{w}_i)^\top (\mathbf{g}_j - \mathbf{g}_i) > 0$, we have $\delta C_{i,j} < 0$.

Now we have proven that both the soft and the hard collisions help reduce the weight correlation, which is correlated to the abstract sum of the weight cosine similarity as indicated in Eq. \ref{eq: wc_def}. The theorem gets proven.
\end{proof}

\begin{proof}[\textbf{Proof of Thm. \ref{thm:convergence}}]
The proposed gradient updates with collision can be generalized in the form that ${\bm\theta}' = \bm\theta - \eta (\bm{g}_{\bm\theta} + \bm\delta)$, where $\bm\theta$ is the model parameters, $\bm\delta$ refers to the introduced collision, $\bm{g}_{\bm\theta}$ represents the original model gradients and $\eta$ is the learning rate. We denote the loss function as $\mathcal{L}(\bm\theta)$, and hence $\bm{g}_{\bm\theta} = \nabla_{\bm\theta} \mathcal{L}$. Suppose $\mathcal{L}(\bm\theta)$ is $L$-smooth, i.e. $\mathcal{L}(\bm\theta') \le \mathcal{L}(\bm\theta) + \bm{g}_{\bm\theta}^\top (\bm\theta' - \bm\theta) + \frac{L}{2} \Vert \bm\theta' - \bm\theta\Vert^2$. If we apply our updates into this inequality, we will get:
\begin{equation}
    \mathcal{L}(\bm\theta') \le \mathcal{L}(\bm\theta) - \eta\bm{g}_{\bm\theta}^\top (\bm{g}_{\bm\theta}+\bm\delta) + \frac{L\eta^2}{2} \Vert \bm{g}_{\bm\theta} + \bm\delta\Vert^2
    \label{eq: loss_theta}
\end{equation}
\textbf{Hard Collision} We will first prove the convergence for hard collision. We summarize several properties of the proposed collision mechanism. Firstly, our collision is a zero-mean collision as indicated from the derivation formula. Hence, we have $\mathbb{E}(\bm\delta_t) = 0$. Secondly, we assume that the gradients are bounded, which is commonly adopted in previous literature~\cite{bottou2018optimization}. Since our proposed collision term is built upon the gradients as shown in Eq. \ref{eq: delta_v}, we can easily prove that $\bm\delta$ is also bounded. Since bounded random variables always have a finite variance, then there exists a $\sigma$ that bounds $\bm\delta$'s variance, i.e. $\mathbb{E}[\Vert\bm{\delta}\Vert^2]\le \sigma^2$.

We would love to investigate the average performance of this stochastic algorithm. We expand Eq. \ref{eq: loss_theta} and take the expectation and get:
\begin{equation}
\begin{aligned}
\mathbb{E}[\mathcal{L}(\bm\theta')] &\le \mathcal{L}(\bm\theta) - \eta (1 - \frac{L\eta}{2})\Vert \bm{g}_{\bm\theta}\Vert^2 - \eta(1 - L\eta) \bm{g}_{\bm\theta}^T \mathbb{E}[\bm\delta] + \frac{L\eta^2}{2} \mathbb{E}[\Vert \bm\delta \Vert^2 ] \\
&\le \mathcal{L}(\bm\theta) - \eta (1 - \frac{L\eta}{2})\Vert \bm{g}_{\bm\theta}\Vert^2 + \frac{L\eta^2}{2} \sigma^2
\end{aligned}
\label{eq: l_sigam}
\end{equation}
We can pick an $\eta_t > 0$ for this iteration that satisfies $\eta_t (1 - \frac{L\eta_t}{2}) > 0$. Next, we want to prove the convergence. We sum Eq. \ref{eq: l_sigam} of various epochs from 1 to T, we have:
\begin{equation}
    \sum_{t=1}^{T} \eta_t (1 - \frac{L\eta_t}{2}) \mathbb{E}[\Vert \bm{g}_{\bm\theta_t} \Vert^2] \le \mathcal{L}(\bm\theta_0) - \mathbb{E}[\mathcal{L}(\bm\theta_{T+1})] + \sum_{t=1}^{T}\frac{L\eta_t^2}{2}\sigma^2
\end{equation}
If we take $\eta_t < \frac{1}{L}$, then $1 - \frac{L\eta_t}{2} > \frac{1}{2}$. Also, since our aim is to minimize $\mathcal{L}(\bm\theta)$, it is reasonable for us to assume that $\mathcal{L}(\bm\theta)$ has a lower bound. Under constant step sizes, we can have that:
\begin{equation}
    \frac{1}{T}\sum_{t=1}^{T}\mathbb{E}[\Vert \bm{g}_{\bm\theta_t} \Vert^2] \le \frac{2}{\eta T}(\mathcal{L}(\bm\theta_0) - \mathbb{E}[\mathcal{L}(\bm\theta_{T+1})]) + L\eta\sigma^2
\end{equation}
If we set $\eta = \frac{\varepsilon}{2L\sigma^2}$, then $L\eta\sigma^2 \le \frac{\varepsilon}{2}$. If we further require $T \ge \frac{4(\mathcal{L}(\bm\theta_0) - \mathbb{E}[\mathcal{L}(\bm\theta_{T+1})])}{\eta\varepsilon} = \mathcal{O}(\frac{\sigma^2}{\varepsilon^2})$, we can achieve $\frac{1}{T}\sum_{t=1}^{T}\mathbb{E}[\Vert \bm{g}_{\bm\theta_t} \Vert^2] \le \varepsilon$.

\textbf{Soft Collision} As for the soft collision, we can prove it by assuming that $\bm\delta$ is uniformly bounded by $D$. This assumption can also be justified by the bound assumption on the gradients since the cosine similarity is bounded by 1. Using this property we have: 
\begin{equation}
    |\bm{g}_{\bm\theta}(\bm\theta)^\top\bm\delta| \le \Vert \bm{g}_{\bm\theta}(\bm\theta)\Vert \Vert \bm\delta \Vert \le \Vert \bm{g}_{\bm\theta}(\bm\theta)\Vert D
\end{equation}
If we insert this into Eq. \ref{eq: loss_theta}, we have:
\begin{equation}
    \mathcal{L}(\bm\theta') \le \mathcal{L}(\bm\theta) - \eta(1 - \frac{L\eta}{2})\Vert \bm{g}_{\bm\theta}\Vert^2 + (L\eta^2 - \eta) \Vert \bm{g}_{\bm\theta}\Vert D + \frac{L\eta^2}{2}D^2
\end{equation}
Then, we want to get rid of the term with solely $\Vert \bm{g}_{\bm\theta}\Vert$. If we apply Young's inequality on this, we get:
\begin{equation}
    \Vert \bm{g}_{\bm\theta}\Vert D \le \frac{\alpha}{2}\Vert \bm{g}_{\bm\theta}\Vert^2 + \frac{1}{2\alpha} D^2
\end{equation}
where $\alpha > 0$ is a constant. For the choice of $\eta$, we can always find an upper bound of it because of our diminishing step-size training scheme. We denote the upper bound as $\hat{\eta}$. Furthermore, if we choose $\eta$s which follow $L\eta - 2 > 0$, we can formulate our inequality as:
\begin{equation}
\begin{aligned}
    \mathcal{L}(\bm\theta') & \le \mathcal{L}(\bm\theta) - \eta(1 - \frac{L\eta}{2})\Vert \bm{g}_{\bm\theta}\Vert^2 + (L\eta^2 - \eta) (\frac{\alpha}{2}\Vert \bm{g}_{\bm\theta}\Vert^2 + \frac{1}{2\alpha} D^2) + \frac{L\eta^2}{2}D^2 \\
    & = \mathcal{L}(\bm\theta) - \eta(1 - \frac{L\eta}{2} + (1 - L\eta)\frac{\alpha}{2})\Vert \bm{g}_{\bm\theta}\Vert^2  + (\frac{L\eta^2}{2} + (L\eta^2 - \eta )\frac{1}{2\alpha})D^2 \\
    & \le \mathcal{L}(\bm\theta) - \eta(1 - \frac{L\hat\eta}{2} + (1 - L\hat\eta)\frac{\alpha}{2})\Vert \bm{g}_{\bm\theta}\Vert^2  + \eta^2(\frac{L}{2} + \frac{L}{2\alpha})D^2 \\ 
    & = \mathcal{L}(\bm\theta) - c\eta\Vert \bm{g}_{\bm\theta}\Vert^2  + c'\eta^2D^2
\end{aligned}
\end{equation}
where $c, c'$ are two constants independent of t. Then, as in the previous case, we sum from epoch 1 to T, we get: 
\begin{equation}
    \frac{1}{T}\sum_{t=1}^{T}\mathbb{E}[\Vert \bm{g}_{\bm\theta_t} \Vert^2] \le \frac{2}{c\eta T}(\mathcal{L}(\bm\theta_0) - \mathbb{E}[\mathcal{L}(\bm\theta_{T+1})]) + \frac{c'\eta D^2}{c}
\end{equation}
Firstly, we choose $\eta=\frac{c\varepsilon }{2c'D^2}$, then $\frac{c'\eta D^2}{c} \le \frac{\varepsilon}{2}$. If we further let $T \ge \frac{4(\mathcal{L}(\bm\theta_0) - \mathbb{E}[\mathcal{L}(\bm\theta_{T+1})])}{c\eta} = \mathcal{O}(\frac{D^2}{\varepsilon^2})$, we can get $\frac{1}{T}\sum_{t=1}^{T}\mathbb{E}[\Vert \bm{g}_{\bm\theta_t} \Vert^2] \le \varepsilon$.

Furthermore, we state that the analysis on the momentum, Adam-like or other types of gradient descent schemes can be directly applied to the ones with the collision mechanism. This is due to the boundedness of the collision term. It will only incur a constant level change on the final result.
\end{proof}

\section{Additional Experiments and Results}
\begin{figure}
    \centering
	\begin{minipage}[b]{0.9\textwidth}
    \centering
		\subfigure[Tanh]{
			\includegraphics[width=0.32\textwidth]{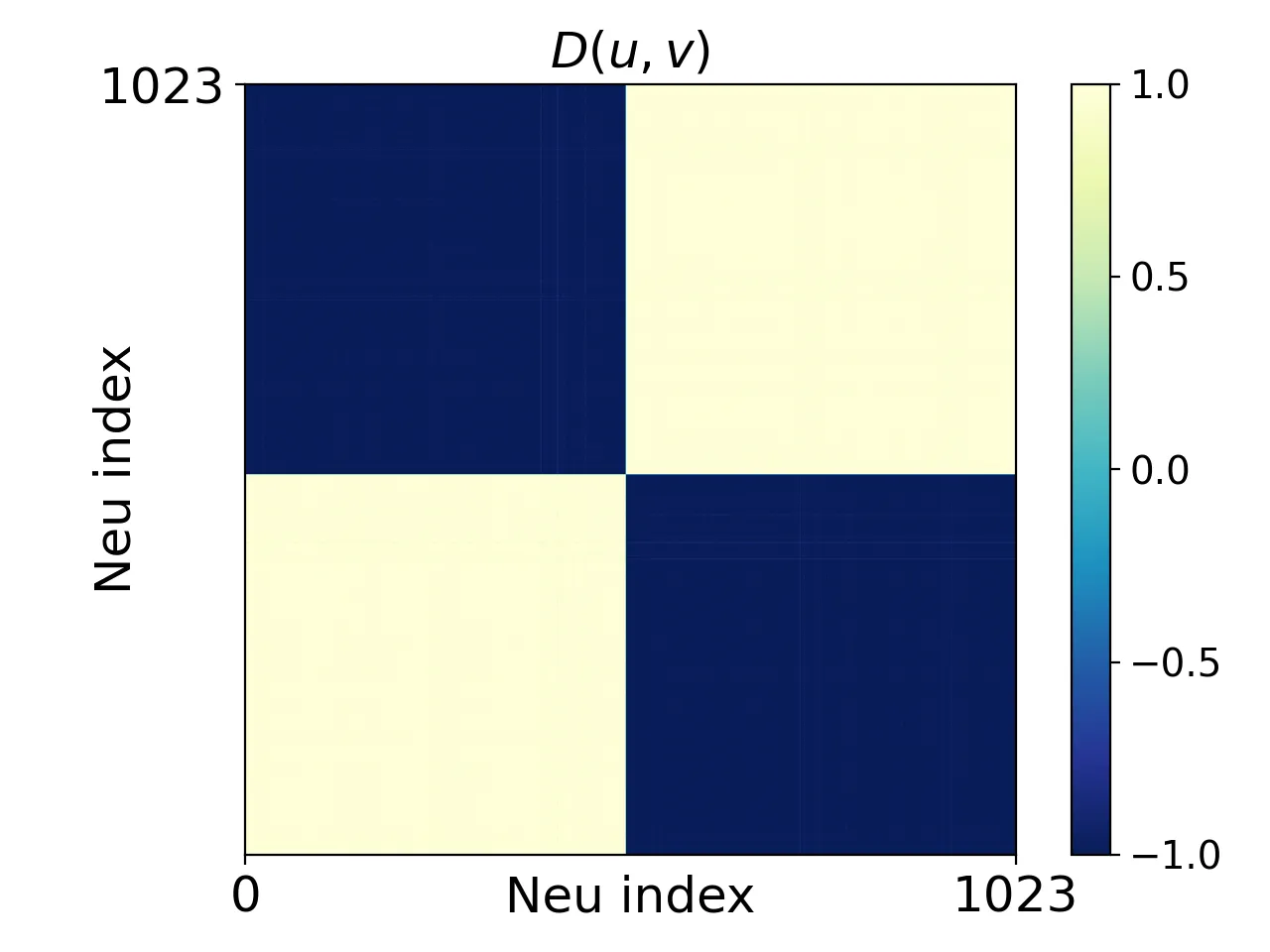} 
			\includegraphics[width=0.32\textwidth]{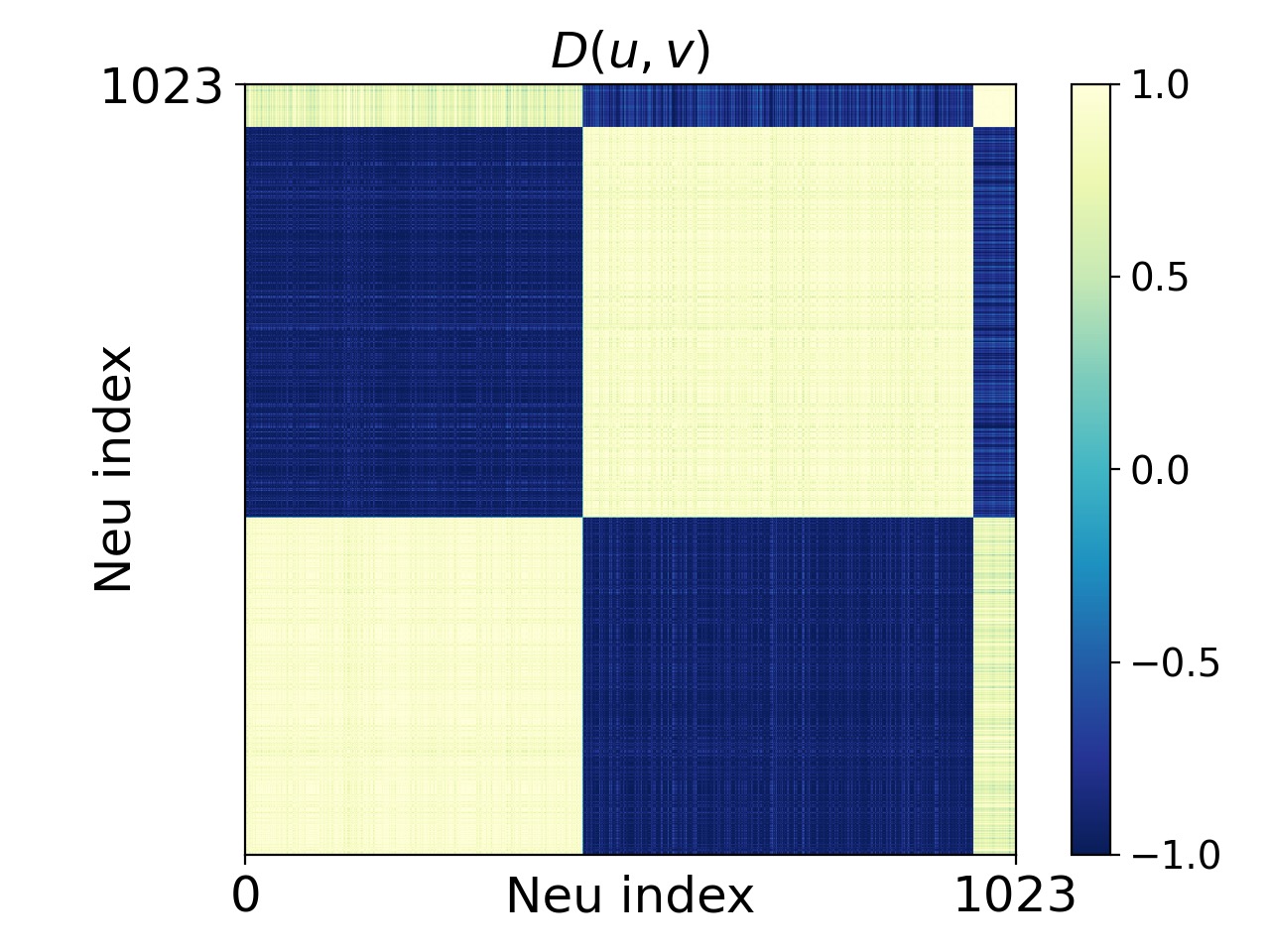}
                \includegraphics[width=0.32\textwidth]{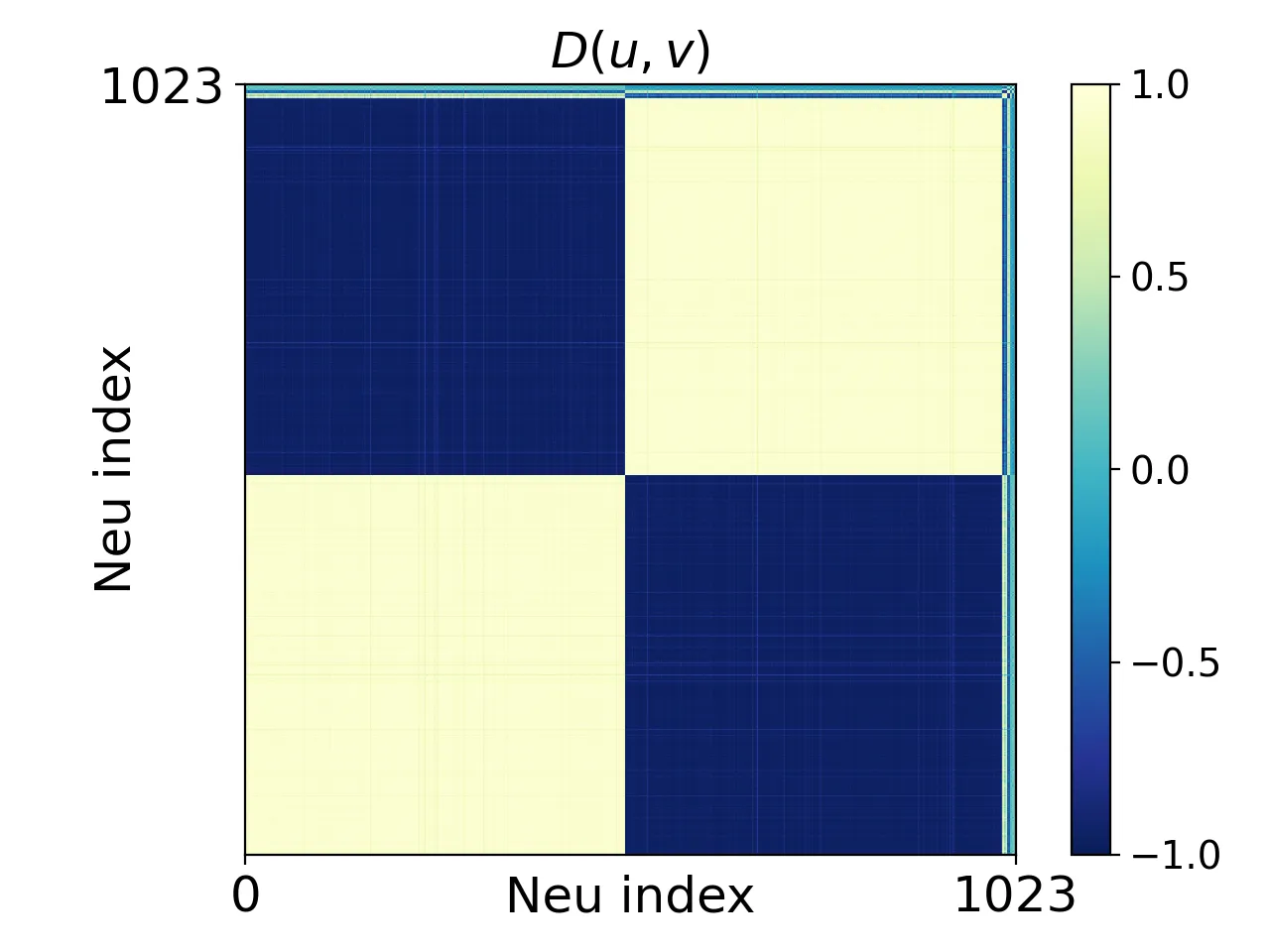}
	}
	\end{minipage}
	\begin{minipage}[b]{0.9\textwidth}
        \centering
		\subfigure[xTanh]{
			\includegraphics[width=0.32\textwidth]{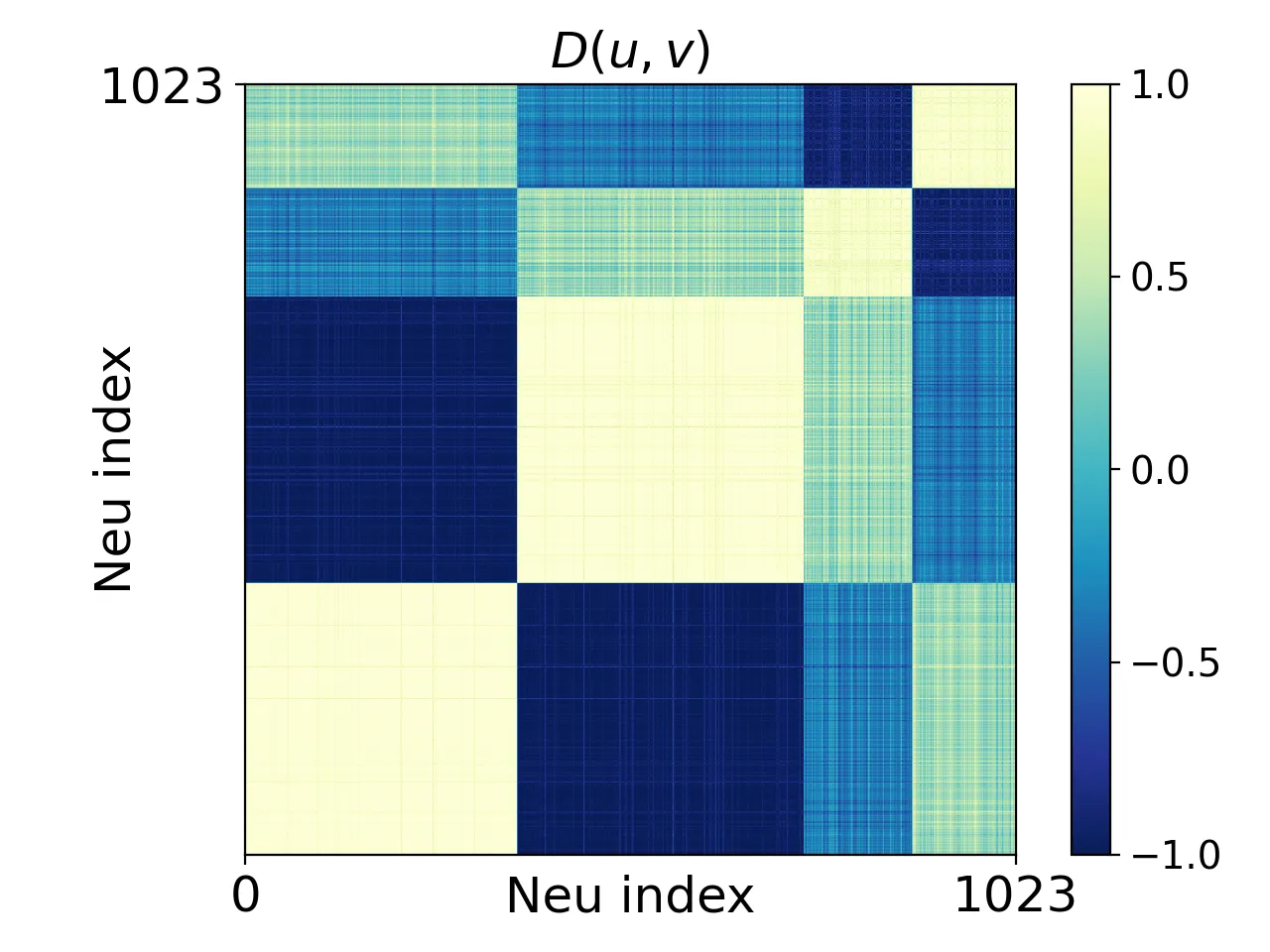} 
			\includegraphics[width=0.32\textwidth]{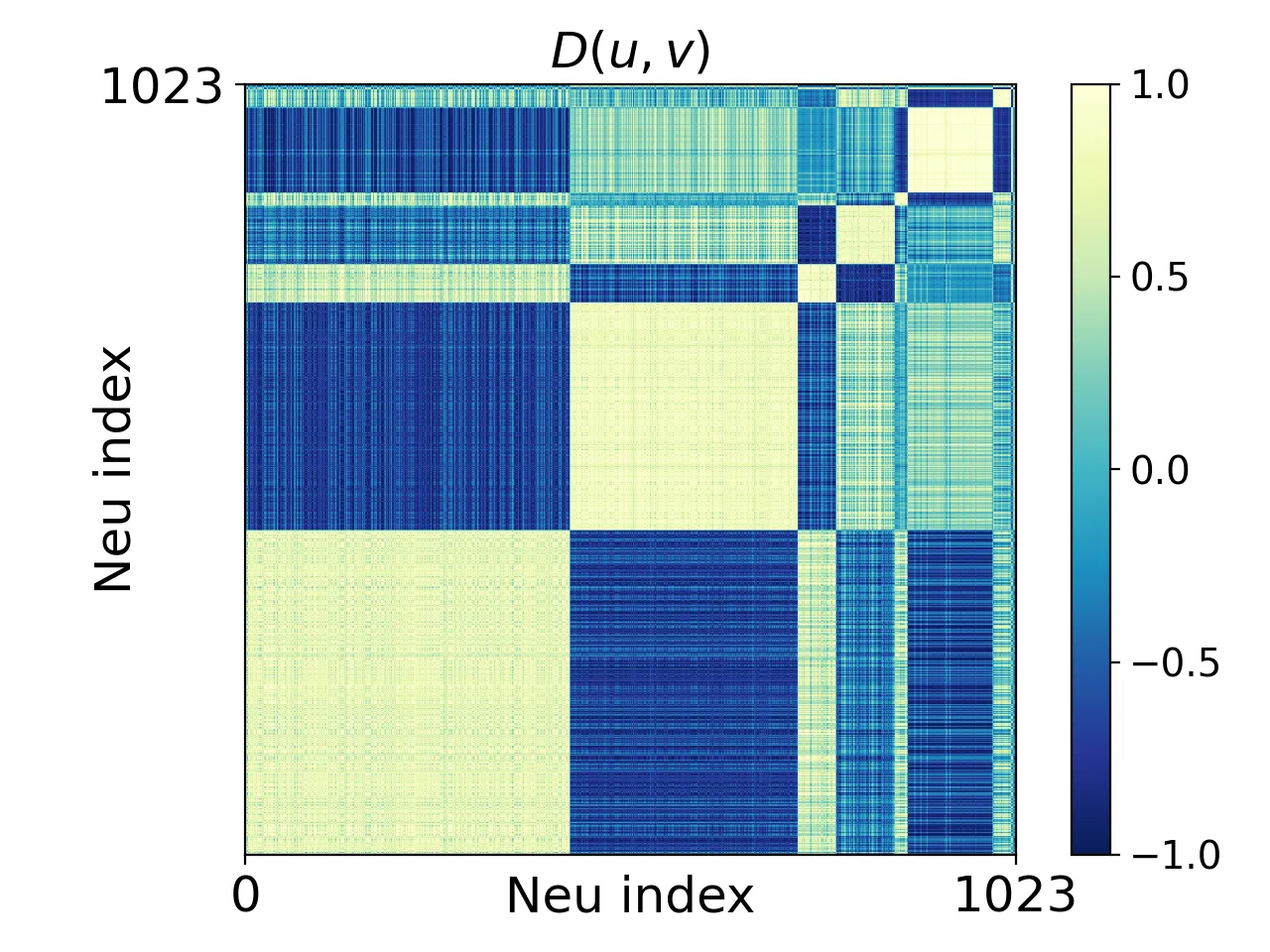}
                \includegraphics[width=0.32\textwidth]{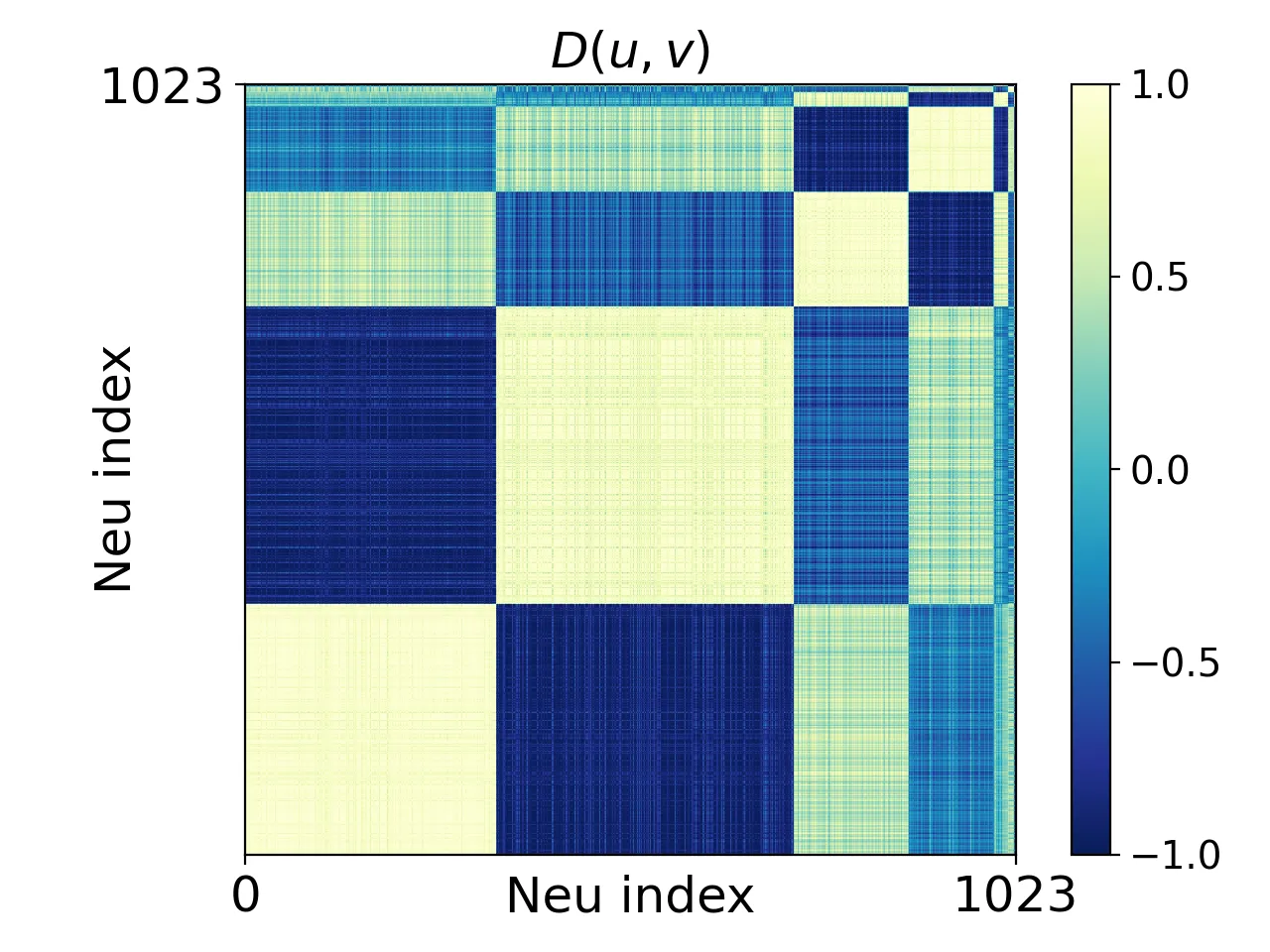}
	}
	\end{minipage}
	\caption{Condensation of Resnet18-like neural networks on CIFAR-10. The color in the figures indicates the cosine similarity of the normalized input weights of two neurons in the first FC layer. The subcaption represents the activation function in the FC layers. The first column shows the model weights trained with the original Adam optimizer. The second and third columns depict the results with Soft Collision and Hard Collision, respectively.}
	\label{fig:resnet}
\end{figure}

\subsection{Condensation Experiments on CIFAR-10.}
\label{appx:condense}
Firstly, we detail the plotting scheme used for condensation visualization. To be specific, after we get the cosine similarity matrix for the weight matrix, we perform clustering based on this similarity. Starting from neuron 0, we iteratively group neurons that are strongly similar (>0.6) to the current neuron. If a negatively correlated neuron (< -0.6) is found, it may become the new reference. After we get these clusters of neurons, we rearrange the cosine similarity matrix based on the new order. The visualized picture represents different clusters of neurons. When the reordered cosine similarity matrix shows large, dense blocks, it indicates that many neurons are highly correlated and thus condensed into a small number of clusters. In contrast, when the similarity matrix exhibits more fragmented or evenly distributed clusters, it implies weaker correlations among neurons. 

Secondly, we test our methods' anti-condensation effects on a more practical example, CIFAR-10~\citep{krizhevsky2009learning}. We follow the ResNet18-like structure introduced in \cite{zhou2022understandingcondensationneuralnetworks}: the original single fully-connected (FC) layer is replaced with a series of FC layers of size 1024-1024-10. The results are visualized in Fig. \ref{fig:resnet}. The learning rate is $3 \times 10^{-8}$ for Tanh activation and $5 \times 10^{-6}$ for xTanh activation. As in the synthetic data experiments, both of the proposed collision methods slow down the original condensation.

\textbf{Weight Decay Ablation.} To verify our claim in the introduction that weight decay does not prevent ultimate condensation, we conduct an ablation on the synthetic dataset (Section~5.1) tracking weight correlation at initialization, mid-training, and end of training under different weight decay strengths:

\begin{table}
\centering
\caption{Experiment results with different weight decay strengths.}
\begin{tabular}{c|c|c|c}
\toprule
Weight Decay & Init $\rho(\mathbf{w})$ & Mid $\rho(\mathbf{w})$ & End $\rho(\mathbf{w})$ \\
\midrule
0        & 0.3867 & 0.9672 & 0.9999 \\
$10^{-3}$ & 0.3867 & 0.9676 & 0.9999 \\
$10^{-1}$ & 0.3867 & 0.9245 & 0.9992 \\
\bottomrule
\end{tabular}
\end{table}

While strong weight decay ($10^{-1}$) transiently reduces mid-training condensation (0.9245 vs.\ 0.9676), all configurations ultimately converge to near-perfect correlation ($\rho \approx 1.0$). This confirms that weight decay, as a norm-based regularizer, cannot provide the angular repulsion needed to maintain diversity at convergence, motivating the need for a mechanism like KO that directly counteracts angular condensation.

\subsection{Additional LLM Experiments}
We further evaluate KO in a chemical-domain LoRA supervised fine-tuning (SFT) setting, adapting Qwen3.5-0.8B-Base and Qwen3.5-2B-Base~\cite{teamqwen3}. Here KO is applied only to the LoRA-B~\cite{hu2022lora} gradients of three Transformer blocks, after distributed gradient synchronization and gradient clipping but before the AdamW update.

The SFT data are drawn from ChemPile~\cite{mirza2026chempile}. We select single-turn, known-topic records spanning seven domains---analytical, general, inorganic, organic, and physical chemistry, together with materials science and technical chemistry. Preprocessing comprises assistant-target expansion, domain filtering, downsampling of the over-represented general-chemistry split, class balancing, and a fixed train/test partition. Tokenization uses the model-specific Qwen3.5 tokenizer with a 2,048-token cutoff and reasoning mode disabled. The baseline and KO runs consume identical records in identical order and share the same effective batch size.

We evaluate on the nine-task ChemBench suite introduced by ChemLLM~\cite{zhang2024chemllm}: Caption2mol, Mol2caption, Name Conversion, Product Prediction, Property Prediction, Retrosynthesis, Solvent Prediction, Temperature Prediction, and Yield Prediction. Of the 4,009 local test records, the 4,008 that fit the 8,192-token prompt limit are scored, with fixed few-shot demonstrations supplied from a separate development split. The ChemBench test records are used for evaluation only, never for SFT or for selecting KO hyperparameters.

Because the baselines were obtained from earlier runs, their hardware configuration differs: the KO runs use eight H200 GPUs with a per-device batch size of 4, whereas the 0.8B and 2B baselines use four and two H200 GPUs with per-device batch sizes of 8 and 16, respectively. Crucially, the effective batch size (32), optimizer steps, learning-rate schedule, training records, data order, and random seed are held fixed across all runs, so the comparison isolates the contribution of KO.

\textbf{Qwen3.5-0.8B-Base.} Table~\ref{tab:qwen0.8B} reports the results. Across repeated runs, the paired task-macro improvement has mean $0.01309$ and standard deviation $0.00407$, indicating a small but consistent gain from KO.

\begin{table}[h]
    \centering
    \caption{Small-scale chemical capability transfer under Qwen3.5-0.8B-Base LoRA SFT.}
    \begin{tabular}{c|c|c|c}
    \toprule
        Treatment & Task-macro & Sample Accuracy & Correct  \\
        \midrule
        Historical Standard LoRA & 0.27653 & 27.87\% & 1,117 / 4,008 \\
        LoRA + KO & 0.28954 & 28.49\% & 1,142 / 4,008 \\
        \midrule
        KO - baseline & +0.01301 & +0.62\% & +25 \\
         \bottomrule
    \end{tabular}
    \label{tab:qwen0.8B}
\end{table}

\textbf{Qwen3.5-2B-Base.} Table~\ref{tab:qwen2B} reports the results. The paired task-macro improvement has mean $0.00864$ and standard deviation $0.00311$, again positive across runs.

\begin{table}[h]
    \centering
    \caption{Small-scale chemical capability transfer under Qwen3.5-2B-Base LoRA SFT.}
    \begin{tabular}{c|c|c|c}
    \toprule
        Treatment & Task-macro & Sample Accuracy & Correct \\
        \midrule
        Historical Standard LoRA & 0.36623 & 35.48\% & 1,422 / 4,008 \\
        LoRA + KO & 0.37491 & 36.40\% & 1,459 / 4,008  \\
        \midrule
        KO - baseline & +0.00867 & +0.92\% & +37 \\
         \bottomrule
    \end{tabular}
    \label{tab:qwen2B}
\end{table}

\textbf{Summary.} Taken together, these results indicate that the benefits of KO extend beyond pretraining and vision to parameter-efficient domain adaptation of LLMs. At both scales, restricting collisions to the LoRA-B gradients of only three Transformer blocks produces a consistent positive shift in chemical task-macro performance ($+0.013$ at 0.8B and $+0.009$ at 2B, corresponding to $+0.62\%$ and $+0.92\%$ sample accuracy), and the paired improvement stays positive across repeated runs at low variance. Although the absolute gains are modest, their consistency across scales, tasks, and seeds---obtained by perturbing only a small fraction of the trainable parameters---suggests that mitigating weight condensation remains beneficial even in the low-rank regime of LoRA fine-tuning.

\subsection{Memory and Runtime}
In this section, we would love to discuss the memory and runtime issues possibly incurred by the collision mechanism. Our experiments on CIFAR datasets are run on NVIDIA 3090 GPUs. The results are shown in Table~\ref{table: cifar10-memory} and Table~\ref {table: cifar100-memory}. We can infer from the table that the soft collision mechanism incurs minimal additional memory consumption. This is because it avoids the construction of large auxiliary matrices, making it suitable even for relatively large models. In contrast, the hard collision method involves the computation of a relative speed matrix, which introduces additional memory overhead. While this cost is manageable for small to medium-scale networks, it may become prohibitive for very large models with a high number of neurons (particles), thus limiting their practical applicability in such settings.

\begin{center}
  \begin{minipage}{0.49\textwidth}
  \captionof{table}{Runtime (in seconds) and memory (in MB) for different ResNet architectures with Vanilla, soft collision, and hard collision on CIFAR-10.}
    \centering
    \begin{tabular}{l|c|c}
\toprule
 & RunTime & Memory \\
\midrule
Vanilla Resnet18 & 5523s  & 2120M \\
Resnet18+S.C.    & 5791s  & 2120M \\
Resnet18+H.C.    & 8784s  & 2124M \\
Vanilla Resnet34 & 5915s  & 2612M \\
Resnet34+S.C.    & 6481s  & 2612M \\
Resnet34+H.C.    & 7293s & 2616M \\
Vanilla Resnet50 & 12039s & 5172M \\
Resnet50+S.C.    & 14871s & 5172M \\
Resnet50+H.C.    & 17334s & 5176M \\
\bottomrule
\end{tabular}
\label{table: cifar10-memory}
  \end{minipage}
  \hfill
  \begin{minipage}{0.49\textwidth}
    \centering
    \captionof{table}{Runtime (in seconds) and memory (in MB) for different ResNet architectures with Vanilla, soft collision, and hard collision on CIFAR-100.}
    \begin{tabular}{l|c|c}
\toprule
 & \makecell{RunTime} & \makecell{memory}\\
\midrule
Vanilla Resnet18 & 5774s & 2122M  \\
Resnet18+S.C & 5802s&2122M\\
Resnet18+H.C &16389s& 2126M\\
Vanilla Resnet34 & 8857s&2612M \\
Resnet34+S.C& 8968s&2612M\\
Resnet34+H.C&18405s &2616M\\
Vanilla Resnet50  & 13308s& 11576M \\
Resnet50+S.C & 13310s& 5176M\\
Resnet50+H.C & 26438s &11576M \\
\bottomrule
\end{tabular}
\label{table: cifar100-memory}
  \end{minipage}
\end{center}

More specifically, suppose that the collision is applied to a linear layer with weight matrix $\mathbf{w} \in \mathbb{R}^{n \times n}$. For soft collision, the algorithm primarily involves matrix multiplications to compute cosine similarity matrices (e.g., $\mathbf{w}\mathbf{w}^\top$) and the repulsion force. For time complexity, the operation is dominated by matrix multiplication, resulting in a theoretical time complexity of $\mathcal{O}(n^3)$. It further requires storing intermediate similarity matrices, hence scaling with space complexity of $\mathcal{O}(n^2)$. Although $\mathcal{O}(n^3)$ seems computationally heavy, it translates to high arithmetic intensity operations (pure matrix multiplications) that are heavily optimized on modern GPUs (e.g., via Tensor Cores). Consequently, the actual runtime overhead is negligible compared to the memory-bound operations of a full training step.

In contrast, the hard collision mechanism simulates physical N-body interactions, which requires broadcasting tensors to compute pairwise relative positions and velocities. This broadcasting creates intermediate tensors of size $\mathcal{O}(n^3)$, causing the space complexity to scale cubically. This makes hard collision a memory-bound process, significantly increasing the risk of memory overflow (OOM) and reducing training speed due to memory bandwidth limitations.

That being said, we emphasize that soft collision not only reduces memory usage but also achieves superior empirical performance. However, we include results for hard collision because, despite its higher memory consumption, it offers a more direct and interpretable connection to the physical dynamics of the collision. In fact, the soft collision mechanism was originally inspired by the hard version, aiming to retain key conceptual benefits while improving efficiency.

\subsection{Impact on Adversarial Perturbations}
In this section, we attempt to investigate KO's further impact on neural networks. We claim that besides boosting performance, our method actually enhances the trained networks' resilience to adversarial perturbations.

We attribute this improved robustness to diverse feature support. Parameter condensation implies that the network relies on a limited, low-rank set of features. An adversary only needs to perturb inputs along these few directions to fool the model. By enforcing feature diversity via collisions, KO ensures that predictions are supported by a broader, more redundant set of signals, making the decision boundary more robust to single-direction perturbations.

We evaluated robustness on CIFAR-10 against two famous adversarial perturbation methods, i.e., Fast Gradient Sign Method (FGSM)~\cite{goodfellow2015explainingharnessingadversarialexamples} and Projected Gradient Descent (PGD)~\cite{madry2019deeplearningmodelsresistant}, without any adversarial training. 

Experiment results are shown in Table~\ref{tab:adv}. When faced with FGSM perturbations, we observe a significant gain (e.g., +6.17\% on ResNet18 with Hard Collision) in accuracy, confirming resistance to single-step perturbations. While PGD effectively breaks all non-adversarially trained models (resulting in near-zero accuracy across the board), KO still maintains a slight edge over the baseline. This confirms that mitigating weight condensation naturally yields a more robust geometric structure in the parameter space, further highlighting KO's applicability.

\begin{table}[htb]
    \centering
    \caption{Experiment results with different adversarial perturbations}
    \begin{tabular}{c|c|c}
    \toprule
         & FGSM~\cite{goodfellow2015explainingharnessingadversarialexamples} & PGD~\cite{madry2019deeplearningmodelsresistant} \\
         \midrule
        ResNet18 & 40.46\% & 0.08\%\\
        ResNet18+S.C. & 43.61\% & \textbf{0.19\%} \\
        ResNet18+H.C. & \textbf{46.63\%} & 0.14\%\\
        ResNet34 & 42.13\% & 0.04\%\\
        ResNet34+S.C. & 44.72\% & \textbf{0.23\%}\\
        ResNet34+H.C. & \textbf{50.08\%} & 0.12\%\\
        ResNet50 & 38.88\% & 0.00\%\\
        ResNet50+S.C. & 39.53\% & 0.00\%\\
        ResNet50+H.C. & \textbf{40.21\%} & 0.00\%\\
        \bottomrule
    \end{tabular}
    \label{tab:adv}
\end{table}

\section{Ablation Studies}
\subsection{Impact of the collision coefficient}
One key hyperparameter in KO is the collision coefficient. We generally base the selection of it via a standard grid search on the validation set. In this section, we attempt to investigate further its impact on performance. For simplicity, we performed the ablation study on ResNet-50 on CIFAR-10 with soft collisions. The results are presented as follows.
\begin{table}[htb]
    \centering
    \caption{Experiment results of ResNet-50 on CIFAR-10 with different collision coefficients}
    \begin{tabular}{c|c|c|c|c|c|c|c|c|c|c}
    \toprule
        $\text{coll}\_\text{coef}$ & 0.0 (Baseline) & 0.1 & 0.2 & 0.3 & 0.4 & 0.5 & 0.6 & 0.7 & 0.8 & 0.9 \\
        \midrule
        Accuracy & 77.38\% & 78.23\% & 78.80\% & 78.30\% & 77.98\% & 78.92\% & 77.52\% & 77.87\% & 78.01\% & 78.16\% \\ 
        Gain &  - & +0.85\% & +1.42\% & +0.92\% & +0.60\% & +1.54\% & +0.14\% & +0.49\% & +0.63\% & +0.78\%\\
         \bottomrule
    \end{tabular}
    \label{tab:abl}
\end{table}

Experimental results show that the proposed KO consistently outperforms the base model under different collision coefficients. The best performance is achieved when we set the collision coefficient to 0.5, achieving a $1.54\%$ performance improvement. This shows the insensitivity of our method to hyperparameters, which grants a more stable application in more complicated settings.

\subsection{Impact of Weight Decay}
The proposed KO is designed as a "plug-and-play" module compatible with standard optimizers. Consequently, we can employ many other optimization improvement techniques to boost the model performance. In this paper, all results reported in our main manuscript already include weight decay as part of the standard training recipe for CIFAR-10/100. To explicitly isolate the contribution of KO versus weight decay, we provide an ablation study below. The results demonstrate three key findings. Firstly, models trained with KO alone consistently outperform those trained with Weight Decay alone. Secondly, when KO is added to a baseline that already uses Weight Decay, performance improves further (e.g., ResNet34 improves from 95.14\% to 95.76\%). Finally, our method functions effectively both with and without weight decay, confirming that KO targets a different aspect of generalization (reducing neuron condensation) than L2 regularization.
\begin{table}[htb]
    \centering
    \caption{Experiment results with and without weight decay with various backbones on CIFAR-10. WD is the abbreviation of weight decay.}
    \begin{tabular}{c|c|c|c|c|c|c}
    \toprule
        & Vanilla & Vanilla + WD & S.C. & S.C. + WD & H.C. & H.C. + WD \\
        \midrule
        ResNet18 & 95.02\% $\pm$ 0.29\% & 95.07\% $\pm$ 0.18\% & 95.74\% $\pm$ 0.09\% & 95.74\% $\pm$ 0.05\% & 95.42\% $\pm$ 0.13\% & 95.58\% $\pm$ 0.07\% \\
        ResNet34 &  95.11\% $\pm$ 0.32\% & 95.14\% $\pm$ 0.37\% &95.52\% $\pm$ 0.12\%& 95.76\% $\pm$ 0.08\% &95.33\% $\pm$ 0.18\%& 95.56\% $\pm$ 0.14\% \\
        ResNet50 & 95.31\% $\pm$ 0.27\% &95.37\% $\pm$ 0.38\% &95.62\% $\pm$ 0.08\%& 95.83\% $\pm$ 0.16\% &95.42\% $\pm$ 0.11\%&95.57\% $\pm$ 0.05\% \\
        \bottomrule
    \end{tabular}
    \label{tab:placeholder}
    \vspace{-10pt}
\end{table}

\section{Clarification of Connections between Physics and Neural Networks}
\label{appx: conn}

\begin{table}[h]
    \centering
    \caption{Mapping between kinetic theory concepts and neural network optimization.}
    \begin{tabular}{c|c|c}
    \toprule
       Concept  &  Kinetic Theory (Physical System) & Neural Network Optimization\\
       \midrule
        Agent & 	A single particle & \makecell{A single neuron \\(weight vector $\mathbf{w}_i$)}\\
        \midrule
        Position & Spatial coordinates of the particle & Weight vector $\mathbf{w}_i \in \mathbb{R}^D$\\
        \midrule
        Velocity & Particle velocity vector & \makecell{(Negative) gradient $-\mathbf{g}_i$\\ (update direction)}\\
        \midrule
        Interaction & Stochastic collisions & \makecell{Gradient modification \\(hard/soft collision terms)}\\
        \midrule
        System Dynamics & Boltzmann Transport Equation & \makecell{Optimization trajectory \\(gradient descent + collision)}\\
        \bottomrule
    \end{tabular}
    \label{tab:conn}
\end{table}

In this section, we further clarify the connection we establish between physics and neural network optimization. Empirically, excessive similarity (condensation) among neurons harms generalization. In kinetic theory, particles maintain separation through collisions. We introduce 'collisions' at the gradient level: if weights represent positions, gradients (negative gradients) represent velocities. When neurons become too similar (particles get too close), our method introduces a repulsive term (simulating a collision) that modifies the 'velocity' of the parameters, pushing them apart to maintain diversity without altering the fundamental loss minimization objective. The explicit mapping between kinetic theory concepts and optimization quantities is presented in Table~\ref{tab:conn} in the main text.

We note several points regarding the scope of the analogy. First, neural network training is not a closed thermodynamic system; energy is not conserved in the physical sense, and the 'collisions' we introduce do not correspond to actual physical interactions. Rather, the kinetic theory framework serves as a source of design principles: (i) collisions should conserve the 'center-of-mass' gradient (ensuring the overall descent direction is preserved), (ii) the repulsion should be isotropic on average (hard collision) or aligned with similarity structure (soft collision), and (iii) the collision probability should increase with relative gradient magnitude (Eq.~\ref{eq: coll_mask}). Second, we treat neurons within the same layer as identical particles in the kinetic theory sense; while neurons may learn distinct functions, they are structurally identical (same dimensionality, same activation) and thus satisfy the identical-particle assumption of kinetic theory in the relevant formal sense. Finally, we emphasize that the DSMC method was chosen over the lattice Boltzmann method because DSMC is better suited to the sparse-particle regime (tens to hundreds of neurons per layer vs.\ Avogadro-scale particle counts) and naturally accommodates the stochastic collision sampling that underlies both our hard and soft collision variants.

\section{Broader Impacts}
This paper presents work whose goal is to further study optimization. There are many potential societal consequences of our work, none of which we feel must be specifically highlighted here.

\end{document}